\documentclass[hidelinks]{sig-alternate}
\usepackage{color}
\usepackage[usenames,dvipsnames]{xcolor}

\makeatletter
\def\@copyrightspace{\relax}
\makeatother

\begin{document}
%
\conferenceinfo{KDD' 15}{August 10-13 2015, Sydney, Australia}

\title{\textcolor{black}{AMP:} a new time-frequency feature extraction method for intermittent time-series data}

%
%
%
%
%

\numberofauthors{5} 
%
\author{
%
%
\alignauthor
Duncan S. Barrack\\
       \affaddr{Horizon Digital Economy Research Institute}\\
       \affaddr{University of Nottingham}\\
       \affaddr{NG7 2TU, UK}\\
       \email{duncan.barrack\\
       @nottingham.ac.uk}
\alignauthor
James Goudling\\
       \affaddr{Horizon Digital Economy Research Institute}\\
       \affaddr{University of Nottingham}\\
       \affaddr{NG7 2TU, UK}\\
       \email{james.goulding\\
       @nottingham.ac.uk}
\alignauthor Keith Hopcraft\\
       \affaddr{School of Mathematical Sciences}\\
       \affaddr{University of Nottingham}\\
       \affaddr{NG7 2RD, UK}\\
       \email{keith.hopcraft\\
       @nottingham.ac.uk}
\and  
\alignauthor Simon Preston\\
       \affaddr{School of Mathematical Sciences}\\
       \affaddr{University of Nottingham}\\
       \affaddr{NG7 2RD, UK}\\
       \email{simon.preston\\
       @nottingham.ac.uk}
\alignauthor Gavin Smith\\
       \affaddr{Horizon Digital Economy Research Institute}\\
       \affaddr{University of Nottingham}\\
       \affaddr{NG7 2TU, UK}\\
       \email{gavin.smith\\
       @nottingham.ac.uk}
}
\date{03 June 2015}

\maketitle
\begin{abstract}
\textcolor{black}{
The characterisation of time-series data via their most salient features is extremely important in a range of machine learning task, not least of all with regards to classification and clustering. While there exist many feature extraction techniques suitable for non-intermittent time-series data, these approaches are not always appropriate for \textit{intermittent} time-series data, where intermittency is characterized by constant values for large periods of time punctuated by sharp and transient increases or decreases in value.}

Motivated by this, we present \textit{aggregation, mode decomposition and projection} (AMP) a feature extraction technique particularly suited to intermittent time-series data which contain time-frequency patterns. For our method all individual time-series within a set are combined to form a non-intermittent \emph{aggregate}. This is decomposed into a set of components which represent the intrinsic time-frequency signals within the data set. \textcolor{black}{ Individual time-series can then be fit to these components to obtain a set of numerical features that represent their intrinsic time-frequency patterns. To demonstrate the effectiveness of AMP, we evaluate against the real word task of clustering intermittent time-series data}. Using synthetically generated data we show that a clustering approach which uses the features derived from AMP significantly outperforms traditional clustering methods. Our technique is further exemplified on a real world data set where AMP can be used to discover groupings of individuals which correspond to real world sub-populations.   
\end{abstract}

\textcolor{white}{.}

\noindent MiLeTS workshop in conjunction with KDD' 15 August 10-13 2015, Sydney, Australia

\section{Introduction}
\noindent Extracting numerical features from time-series data is desirable for a number of reasons including revealing human interpretable characteristics of the data \cite{xing2011extracting}, data compression \cite{fu2011review} as well as clustering and classification \cite{garrett2003comparison, liao2005clustering, wang2006characteristic}. It is often useful to divide the different feature extraction approaches into frequency domain and time domain based methods. Frequency domain extraction techniques include the discrete Fourier transform \cite{vlachos2005periodicity, vlachos2006structural} and wavelet transform \cite{misiti2007clustering}. Examples of time domain techniques are model based approaches \cite{macdonald1997hidden} and more recently shapelets \cite{mueen2012clustering, ye2009time}.

Of particular interest to this paper are intermittent time-series data, such as that derived from human behavioural (inter-) actions, e.g. communications and retail transaction logs. Data of this type contains oscillatory time-frequency patterns corresponding to human behavioural patterns such as the 24 hour circadian rhythm, or 7 day working \newline week/weekend. It is also characterised by short periods of high activity followed by long periods of inactivity (intermittence) \cite{barabasi2005origin, jo2012circadian, karsai2012universal, vazquez2006modeling}. Such characteristics mean that intermittent time-series feature sharp transitions in the dependent variable. When frequency based feature extraction techniques underpinned by the Fourier or wavelet transforms are applied, the transforms produce \emph{ringing} artefacts (a well known example in Fourier analysis is the Gibbs phenomena \cite{gibbs1899fourier}) which results in spurious signals being produced in the spectra. These rogue signals make it extremely difficult to determine what the genuine frequency patterns in the data are. Furthermore, such signals are extremely damaging to clustering and classification techniques which use frequency or time-frequency features as inputs. 

\textcolor{black}{
Another feature of intermittency is that it results in time-series that take a single, constant value for very large portions of the time domain. This phenomena severely degrades the effectiveness of using time domain based extraction methods for machine learning tasks. We will demonstrate not only the impact intermittent data has on traditional extraction methods (showing that the more intermittent the data, the greater the deleterious impact of this effect) but go on to present a new solution to this issue.}

The paper is structured as follows. In section \ref{sec:prob statement} we demonstrate the issue of using traditional feature extraction techniques on intermittent data with a focus on the use of derived features for clustering. After discussing related work in Section \ref{sec:related_work} we introduce our ameliorative strategy in the form of Aggregation, mode decomposition and projection (AMP) in Section \ref{sec:method}. We show in Section \ref{sec:eval} that when features derived from AMP are used for clustering synthetically generated intermittent time-series data, results are significantly better than those which use traditional time-series clustering techniques. In this section, we also demonstrate that AMP gives promising results when applied a real communications data set. We conclude with a discussion in Section \ref{sec:conclusions}. 

\section{Background}
\label{sec:prob statement}
\begin{figure}[htbp]
\centering
\includegraphics[width=0.5\textwidth]{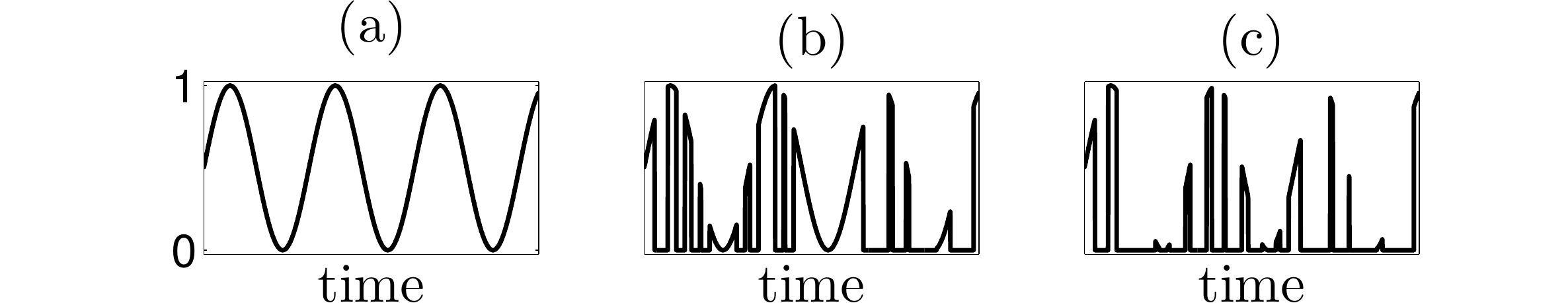}
\caption{\small Three example time-series illustrating the distinctions between a non-intermittent times series (a), partially intermittent time-series (b) and an extremely intermittent time-series (c).}
\label{fig:time_series_IM_egs}
\end{figure}

\noindent Although the concept of intermittence has received some examination across various fields \cite{scott2013encyclopedia, kicsi2009neural} no accepted definition for the term currently exists. Consequently, in this work, we introduce our own expression which can be used to quantify intermittence in time-series. Before this is formally defined, to explain our rationale behind it we refer the reader to plots of three time different time-series in Figure \ref{fig:time_series_IM_egs}, \textcolor{black}{ all with the same frequency pattern. Clearly time-series (a) is non-intermittent, with (b) being somewhat intermittent and (c) extremely intermittent - it takes a value of zero for large portions of the time domain, and consequently its frequency pattern is much harder to identify.}

These observations leads us to construct a \textcolor{black}{practical} intermittence measure based on the total proportion of the time domain that a time-series takes its most frequent value. In particular, if we regard a discrete time-series as a vector $\mathbf{x}=(x_{1},x_{2},\ldots,x_{n})$ of real valued elements sampled at equal spaces in time as a realisation of a random process, then we define a measure of its intermittence by 
\begin{align}
\phi (\mathbf{x}) = P(x_{i}=M(\mathbf{x})),
\label{eq:interittence}
\end{align}
\noindent where $M(\mathbf{x})$ is the mode, or most common value, in vector $\mathbf{x}$ and $P(x_{i}=M(\mathbf{x}))$ is the empirical probability, or relative frequency, that a randomly selected element $x_{i}$ of $\mathbf{x}$ has this value. As a time-series becomes increasing intermittent $\phi$ will tend to 1. With this definition, time-series (a) in Figure \ref{fig:time_series_IM_egs} has a value for $\phi$ of 0.0001 reflecting the fact it is not intermittent. The intermittence measure of time-series (c) (0.7675) is higher than time-series (b) (0.4860) reflecting our observation that (c) is more intermittent than (b).

\begin{figure}[htbp]\hspace{0.7in}time-series \hspace{0.71in}MDS scatter plots \\
\includegraphics[width=0.025\textwidth]{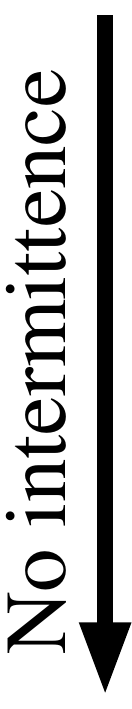}
\includegraphics[width=0.22\textwidth]{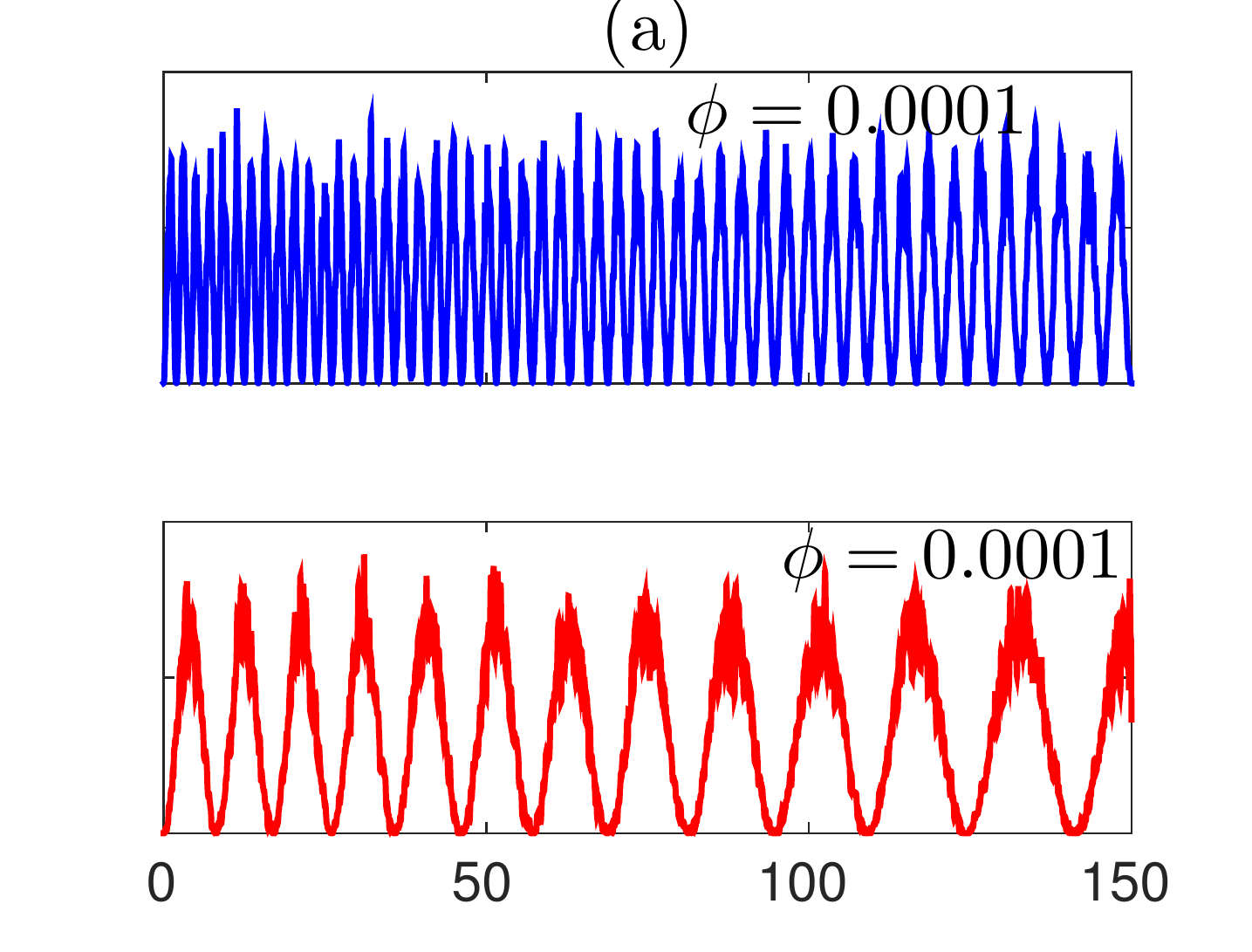}
\includegraphics[width=0.22\textwidth]{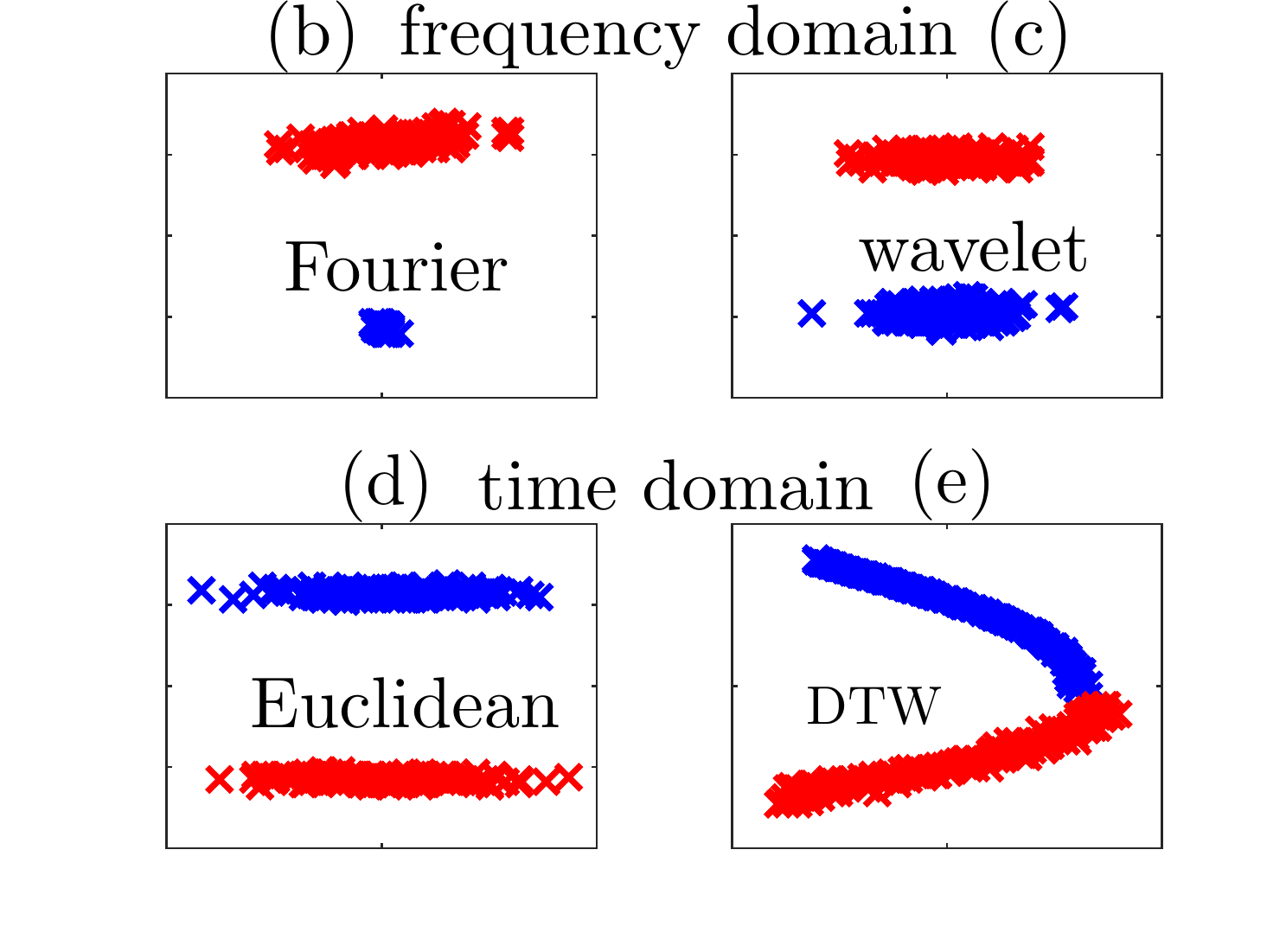}

\includegraphics[width=0.025\textwidth]{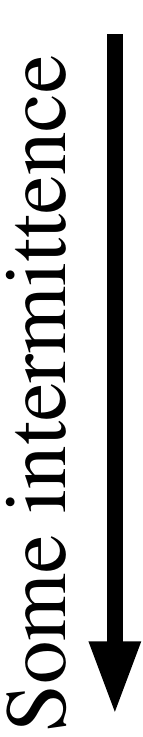}
\includegraphics[width=0.22\textwidth]{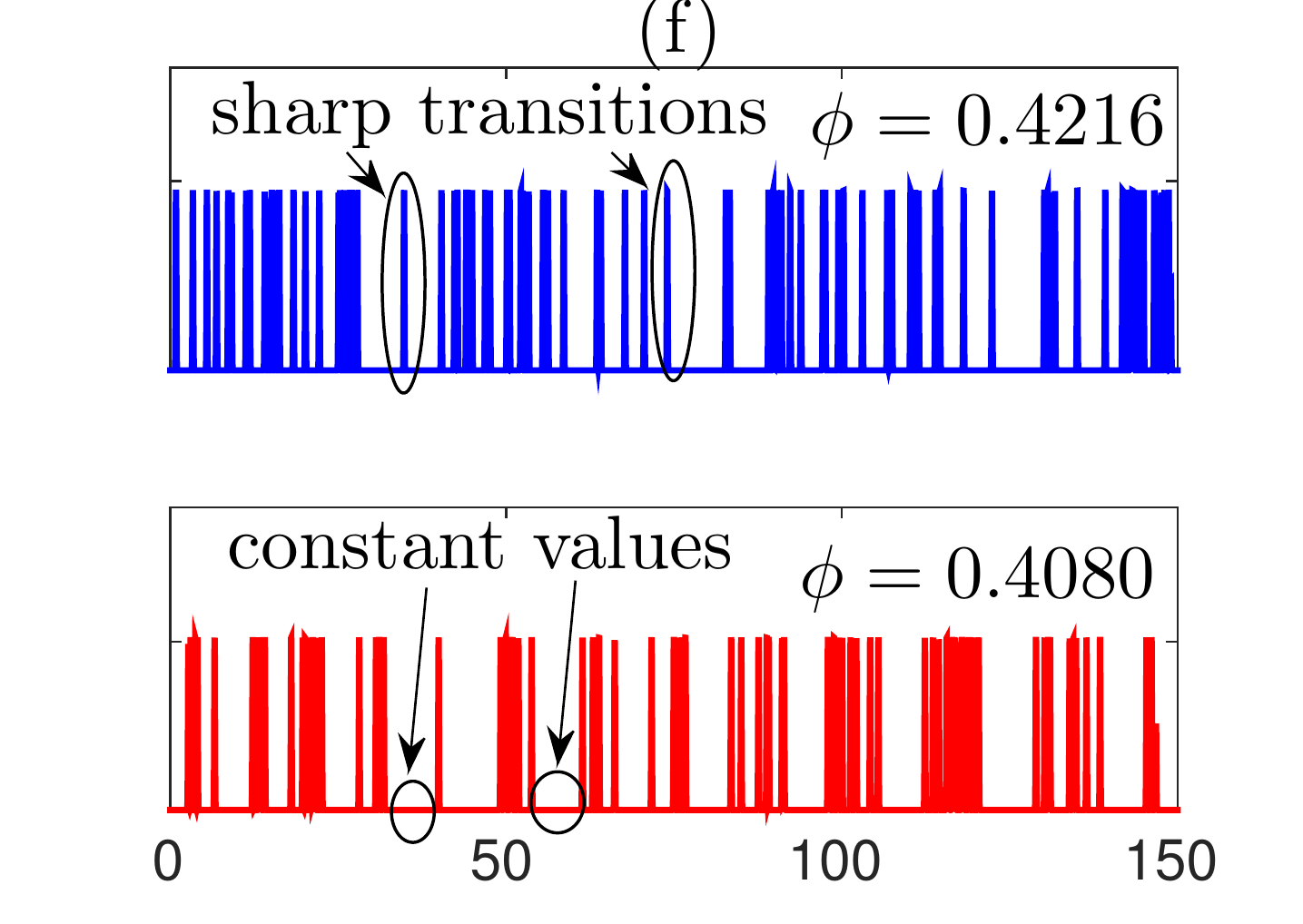}
\includegraphics[width=0.22\textwidth]{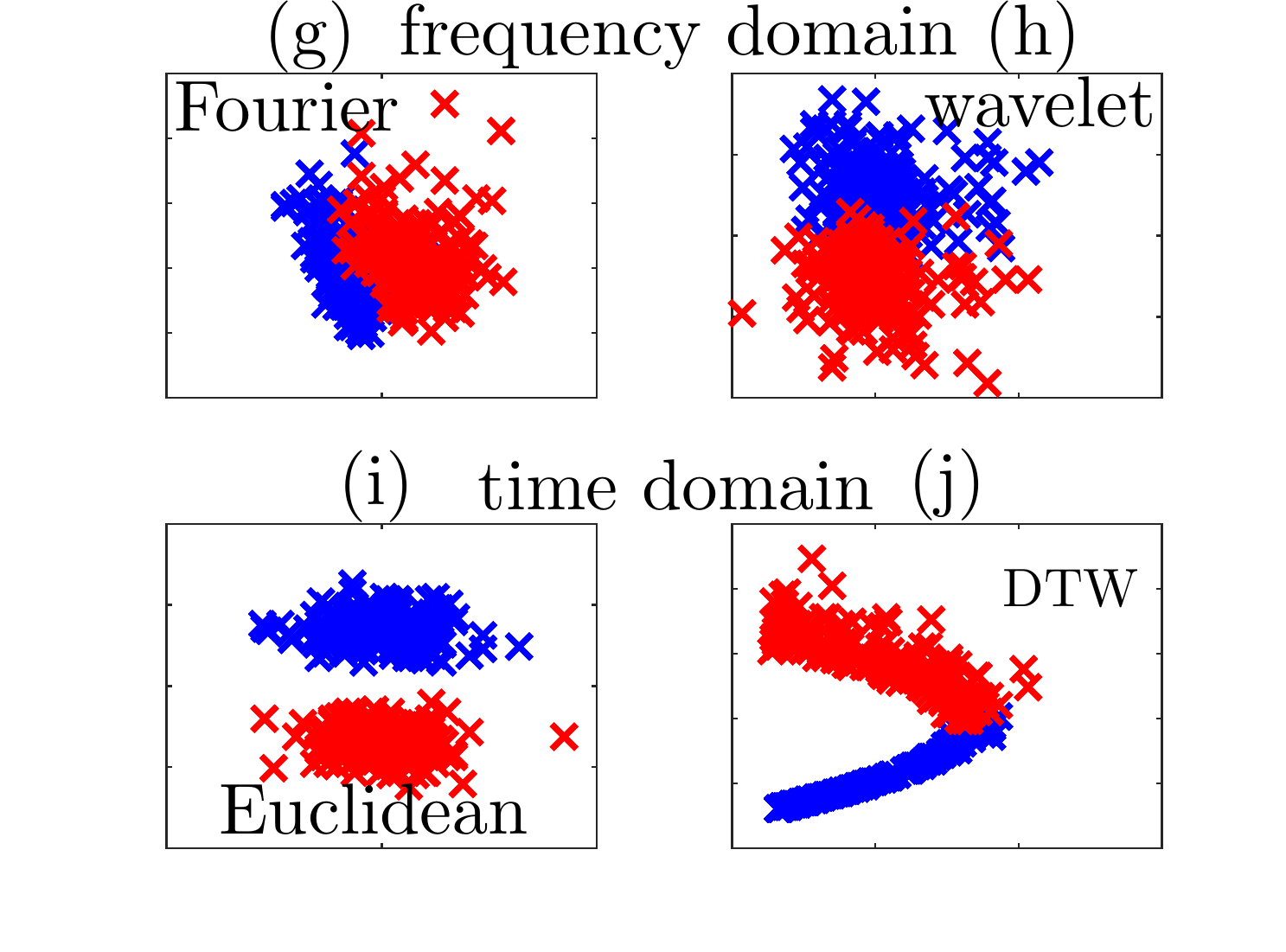}

\includegraphics[width=0.025\textwidth]{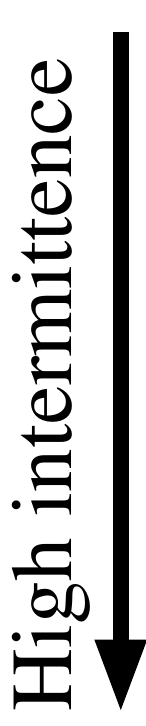}
\includegraphics[width=0.22\textwidth]{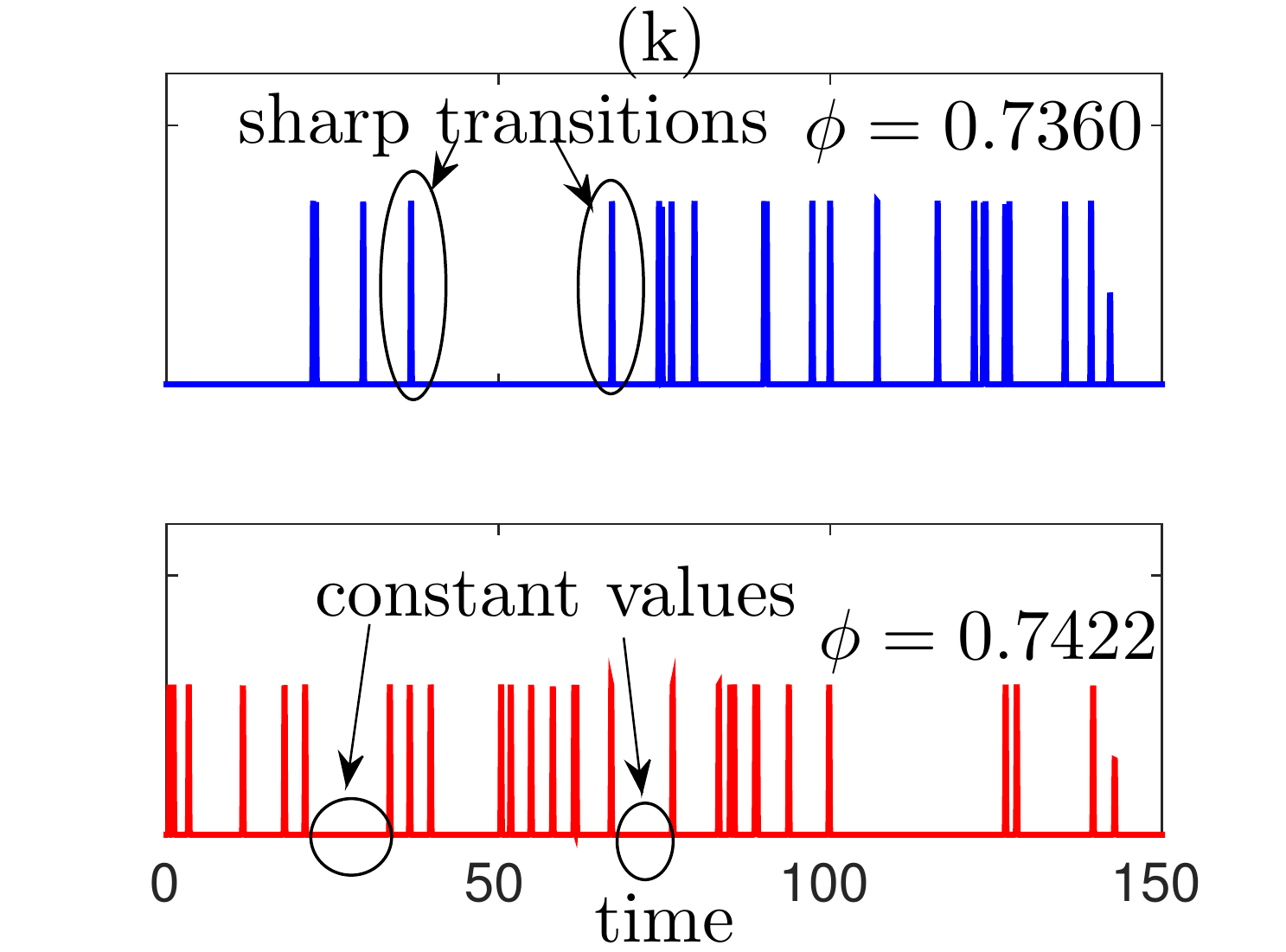}
\includegraphics[width=0.22\textwidth]{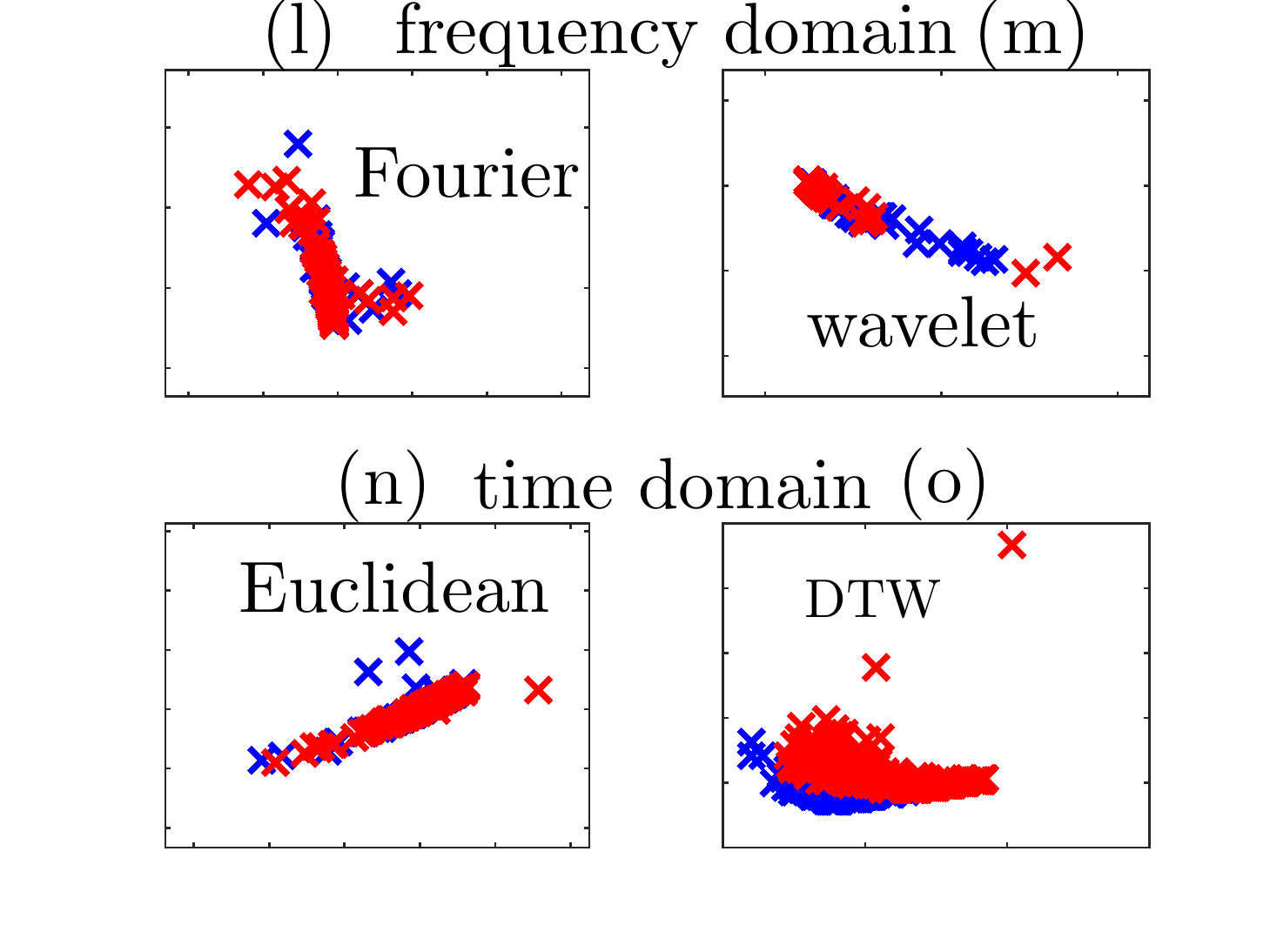}

 \caption{\small The impact of intermittency on cluster analysis. Plot (a) shows two non-intermittent time-series from a set of 200 which were generated via an almost periodically driven stochastic process, with periods ranging linearly from 2 to 4 for one half the set (blue) and 8 to 16 for the other half (red). Each time-series in the set of 200 is plotted using the values for the first two dimensions obtained from classical MDS of every coefficient value of the terms in their direct discrete Fourier (plot (b)) and wavelet decomposition (plot (c)) as well as of the Euclidean (plot (d)) and DTW (plot (e)) distance matrices. Plot (f) shows time-series with increased intermittency but with the same time-frequency pattern as in (a). The corresponding MDS plots are shown in plots (g-j). Finally, plot (k) shows highly intermittent time-series with associated MDS scatter plots in (l-o) illustrating the collapse in efficacy of cluster analysis. The value for the intermittence $\phi$ (equation (\ref{eq:interittence})) for the time-series are given in the figure insets.}
 \label{fig:intermittency_examples} 
 \end{figure}

To illustrate the negative impact that intermittence has on the pertinence of features extracted using traditional techniques, \textcolor{black}{let us} first consider two distinct sets of non-intermittent time-series. We investigate what affect increasing intermittence has on clustering results which use features obtained via the Fourier and wavelet transforms as well as clustering approaches which use the Euclidean and dynamic time warping (DTW) distance between individual time series. The first set of time-series data is composed of 100 realisations of an {\it almost periodically-driven} stochastic process \cite{bezandry2011almost} (see Section \ref{sec:syn_generation} for full details of this procedure), with period ranging linearly from 2 at the beginning of the simulation to 4 at the end (time-series from this set are depicted diagrammatically in blue). The second set also contains 100 time-series generated from an identical process, except for a period which ranges linearly from 8 to 16 (depicted diagrammatically in red). Two examples from each set are illustrated in Figure \ref{fig:intermittency_examples}a. Each of the time-series is plotted using the values for the first two dimensions obtained from classical multi-dimensional scaling (MDS) \cite{cox2010multidimensional} of every coefficient value of each term in their direct discrete Fourier and wavelet decompositions (see Section \ref{sec:synth_eval} for full details of this procedure) in Figures \ref{fig:intermittency_examples}b and \ref{fig:intermittency_examples}c respectively. Additionally MDS results are shown where the Euclidean distance and DTW distance are used as the similarity measure between time series (Figures \ref{fig:intermittency_examples}d and \ref{fig:intermittency_examples}e respectively). Clearly, in this instance, a clustering approach based on any of these techniques is sufficient to discriminate the time-series from the two groups.

Next we consider what effect time-series with a greater value of $\phi$ (and hence higher intermittence) has upon clustering.  These have the same time-frequency patterns as the corresponding time-series in Figure \ref{fig:intermittency_examples}a but are more intermittent. Examples are presented in Figure \ref{fig:intermittency_examples}f and illustrate the sharp transitions and long periods for which the time-series take a constant value (some examples marked in the figure) that begin to occur in the data. Although it is still possible to discriminate the time-series in the MDS plots (Figures \ref{fig:intermittency_examples}g-j), the sharp transitions in the data introduce ringing artefacts in the frequency based decompositions which results in less well separated clusters (compare Figure \ref{fig:intermittency_examples}g with \ref{fig:intermittency_examples}b and Figure \ref{fig:intermittency_examples}h with \ref{fig:intermittency_examples}c). Furthermore, the large periods of constant values act to degrade the discriminative power of Euclidean and DTW based methods (compare Figure \ref{fig:intermittency_examples}i with \ref{fig:intermittency_examples}d and Figure \ref{fig:intermittency_examples}j with \ref{fig:intermittency_examples}e). 

By the time we have increased intermittency further still to generate sets of 100 highly intermittent time-series (Figure \ref{fig:intermittency_examples}k) the negative impact of intermittency on clustering is severe and neither frequency domain based, Euclidean or DTW based methods can be used to separate the data (see Figure \ref{fig:intermittency_examples}l-o).

\section{Related work}
\label{sec:related_work}
\noindent Numerous techniques for \emph{non-intermittent} time-series feature extraction, both time and frequency domain based, have been proposed. The most prevalent use of these within the machine learning community is to obtain numerical features for use as inputs for clustering and classification algorithms.

The most simple time domain feature extraction techniques involve extracting summary statistics such as the mean, variance, as well as other higher order moments of the time-series data. Such features have been used for time-series classification \cite{wang2006characteristic}. Other more complex time-domain features such as the Lyapunov exponent \cite{wolf1985determining} have also been used for machine learning \cite{schreiber1997classification}. Recently, shapelets which represent local features in the data, have been used for classification \cite{ye2009time} and unsupervised learning \cite{mueen2012clustering} with promising results. Model based approaches, where time-series data is fitted to a statistical model are also common. For example the linear predictive coding  cepstrum coefficients obtained from fitting data to the autoregressive integrated moving average model have been used for clustering \cite{kalpakis2001distance}.  
Although not strictly based on a feature extraction technique, an effective approach with regards to time series learning is to use the raw, un-transformed time series data itself. It has been known for some time that using the Euclidean distance as a similarity measure between time-series data can lead to extremely good clustering results \cite{keogh2003need}. Elastic measures including DTW and edit distance, where the temporal alignment of data points isn't respected, are also popular. An empirical study conducted on the data contained within the UCR time-series data mining archive \cite{keogh2002ucr} where the performance of numerous static and elastic measures on classification was investigated suggested that DTW distance is the best measure \cite{ding2008querying}.

Frequency domain based approaches are most commonly underpinned by the discrete Fourier or wavelet transformation of the data. For example Vlachos \emph{et al} used periodic features obtained partly via the direct Fourier decomposition for clustering of MSN query log and electrocardiography time-series data \cite{vlachos2005periodicity}. Features derived from wavelet representations have also been used to cluster synthetic and electrical signals \cite{misiti2007clustering}.  

These works show that time domain and frequency domain based features and distance measures can be extremely effective inputs to classification and clustering algorithms when time-series are non-intermittent. However, we not aware of any work which investigates how these approaches stand up against {\it intermittent} data or present feature extraction approaches designed specifically for data of this type.

\section{Aggregation, Mode Decomposition and Projection (AMP)}
\label{sec:method}
\noindent As a way of dealing with the issues discussed in Section \ref{sec:prob statement}, in this section, we outline our time-frequency feature extraction method. Firstly, all intermittent time-series are pooled into a non-intermittent aggregate. A set of vectors corresponding to pertinent time-frequency pattens is then learnt from the aggregate. By projecting the individual time-series data onto this set, we obtain a set of fitted coefficient values. These act as a feature vector indicating the degree to which each time-frequency pattern of the aggregate is expressed in each individual time-series. The values of the features are suitable for further analysis, e.g. to cluster or classify intermittent time-series data. Such an approach makes three main assumptions (1) a non-intermittent time-series can be obtained from the aggregation of a set of intermittent time-series; (2) the decomposition of the aggregate contains components which correspond only to the underlying time-frequency patterns of the data and not to spurious signals; and (3) the time-frequency patterns of the intermittent data (which due to intermittency are difficult to identify directly) are represented in the aggregate (which because of it non-intermittent nature are far easier to identify directly). The stages of the AMP method are described below.\\

\newpage
\noindent \textbf{Aggregate time-series generation.}
Given a set of $m$ discrete time-series $\mathbf{X} = \{\mathbf{x_{1}}, \mathbf{x_{2}}, \ldots , \mathbf{x_{m}}\}$, where each time-series $\mathbf{x_{i}}=(x_{i1}, x_{i2}, \cdots x_{in})$ is represented by a vector of length $n$ with real valued elements, an aggregate is constructed as follows

\begin{equation}
\mathbf{a}=\sum\limits_{i=1}^{m} \mathbf{x_{i}}.
\label{eq:aggregate_time}
\end{equation}

Under the assumption that each time-series $i$ is a realisation of a stochastic process with a positive probability that it will take a value other than the modal value, as $m \to \infty$, equation (\ref{eq:aggregate_time}) will yield an \emph{non-intermittent} aggregate.  There is evidence to support the notion that much data can be regarded as realisations of stochastic processes with a positive probability that an event will take place at any time. For example the times at which emails are sent has been modelled with a cascading non-homogeneous Poisson process with a positive rate function \cite{malmgren2008poissonian}.  This model led to results with characteristics which were consistent with the characteristics of empirical data.

\noindent \textbf{Time-frequency feature learning.}
Using a signal decomposition technique (e.g. Fourier or wavelet decomposition) $\mathbf{a}$ is decomposed into $l$ components

\begin{equation}
\mathbf{a}=\sum\limits_{j=1}^l \mathbf{b_{j}},
\label{eq:decompose_aggregate}
\end{equation}

\noindent where each vector $\mathbf{b_{j}}=(b_{j1}, b_{j2}, \cdots, b_{jn})$ corresponds to a different time-frequency component. To ensure that (\ref{eq:decompose_aggregate}) does not include a constant term corresponding to the mean of the signal and only includes components corresponding to time-frequency patterns, $\mathbf{a}$ is mean centred (also know as `average centring') prior to its decomposition. 
Series (\ref{eq:decompose_aggregate}) is ordered in descending order of the total energy of each component signal, i.e. $\sum\limits_{k=1}^{n} \mid b_{1k} \mid ^{2} \geq \sum\limits_{k=1}^{n} \mid b_{2k} \mid ^{2} \geq \cdots \geq \sum\limits_{k=1}^{n} \mid b_{lk} \mid ^{2}$. We discard time-frequency components that are not, or only minimally, expressed in the aggregate (these correspond to signals with the lowest energies) as such terms are often an artifact of noise in the data or the decomposition process itself. This is achieved by selecting the first $p$ terms of (\ref{eq:decompose_aggregate}) (where $p \leq l$) \\

\noindent such that $\left(\sum\limits_{j=1}^p \sum\limits_{k=1}^{n} \mid b_{jk} \mid ^{2} \right) \Big{/} \sum\limits_{k=1}^{n} \mid a_{k} \mid ^{2} \geq E_{t}$,\\
where $E_{t} \in [0,1]$ represents a selected threshold. This procedure ensures that only the components which correspond to the most salient time-frequency patterns of the aggregate are selected. Otherwise, the inclusion of terms corresponding to low energy time-frequency patterns in the subsequent projection step of AMP will result in the fitting of intermittent time-series to these unimportant patterns. In this work we set $E_{t}=0.9$, as we find such a value is sufficient to omit low energy signals.

Next, each retained component of (\ref{eq:decompose_aggregate}) is normalised (i.e. $\hat{\mathbf{b_{j}}}=\mathbf{b_{j}}/\mid \mathbf{b_{j} \mid}$). This step is key as it ensures that, during the next step of our method, where each individual time-series is projected onto a set of basis vectors made up of the retained components, each basis vector will have equal weight. This ensures that basis vectors corresponding to components with extreme amplitudes will not skew results in the projection step.

\noindent \textbf{Basis vector projection.}
The final step in our method is to obtain a set of numerical features for each time-series $\mathbf{x_{i}}$ which indicate how much each time-frequency feature learnt from the aggregate is present in them. This is achieved by projecting each $\mathbf{x_{i}}$ on to the set of normalised basis vectors. In particular we seek the 
linear combination of basis vectors which is closest in the least-squares sense to the original observation, i.e. we minimise
\begin{equation}
\mid \mid \mathbf{x}{_{i}}^{\text{T}}-\mathbf{\widehat{B}}\boldsymbol{c}{ _{i}}^{\text{T}} \mid \mid
\label{eq:matrix}
\end{equation}
\noindent where the $n \times p$ matrix $\mathbf{\widehat{B}} = (\hat{\mathbf{b}}_1^{\text{T}}, \hat{\mathbf{b}}_2^{\text{T}}, \ldots, \hat{\mathbf{b}}_p^{\text{T}})$ is comprised of normalised basis vectors learnt from the aggregate. $\mathbf{c_{i}}=(c_{i1}, c_{i2}, \ldots , c_{ip})$ is a vector of fitted coefficients which form the feature vector.  The value of element $c_{ij}$ indicates the degree to which the time-frequency signal corresponding to normalised basis vector $j$ is expressed in time-series $i$.

Fitting all $m$ time-series to the set basis of vectors, as described above, yields the set of features \{$\boldsymbol{c}_{1}$, $\boldsymbol{c}_{2}$, $\boldsymbol{c}_{3}$,\ldots ,$\boldsymbol{c}_{m}$\}. This feature set therefore represents the extent to which an individual time-series expresses the time-frequency patterns present within the overall population. Clustering on this set will result in the grouping together of time-series with similar time-frequency patterns and the clustering into different groups of those which exhibit different time-frequency patterns.  \\

\noindent \textbf{Choice of decomposition method for the aggregate time-series.}
We consider four methods for the decomposition of the aggregate $\mathbf{a}$, which were selected based on the high prevalence in which they appear in the signal processing literature. These are described below. 

\noindent {\bf Discrete Fourier decomposition.} Using the discrete Fourier transform (DFT) \cite{oppenheim1989discrete} the aggregate is decomposed into a Fourier series. We set the number of Fourier components $l=1022$. This ensures that the Fourier series approximates the aggregate extremely well for all data considered in this paper whilst, at the same time, being relatively computationally inexpensive to obtain. We refer to the variant of AMP which uses Fourier decomposition for the aggregate as discrete Fourier transform AMP (DFT-AMP). 

\noindent {\bf Discrete wavelet decomposition.} This decomposition procedure takes a wavelet function and decomposes a time-series in terms of a set of scaled (stretched and compressed) and translated versions of this function \cite{sheng2000wavelet}. Because of it's prevalence of use within the scientific literature we use the Haar wavelet \cite{mallat1999wavelet} for the mother wavelet. For consistency with DFT-AMP we ensure that the discrete wavelet transform produces 1022 components. This approach as discrete wavelet transform AMP (DWT-AMP).

\noindent {\bf Discrete wavelet packet decomposition.} The wavelet packet transform \cite{oppenheim1995wavelets} is a generalisation of the wavelet transform which provides a more flexible data adaptive decomposition of a signal. It can be used to produce a sparser representation and consequently it is preferred to the wavelet transform when signal compression is the goal. Unlike the DWT there is no fixed relationship between the number of basis functions at each scale. The set of wavelet packet basis functions is selected according to the minimisation of a cost function. We again use the Haar wavelet and select the optimal basis set using the Shannon entropy criteria for the cost function \cite{coifman1992entropy}. The variant of AMP which uses DWPT is referred to as discrete wavelet packet transform AMP (DWPT-AMP).

\noindent {\bf Empirical Mode Decomposition.} In contrast to Fourier and wavelet decomposition, empirical mode decomposition (EMD) \cite{huang1998empirical} makes no \emph{a priori} assumptions about the composition of the time-series signal and as such is completely non-parametric. The method proceeds by calculating the envelope of the signal via spline interpolation of its maxima and minima. The mean of this envelope corresponds to the intrinsic mode of the signal with the highest frequency and it is designated the first {\it intrinsic mode function} (IMF). The first IMF is then removed from the signal and lower frequency IMFs are found by iteratively applying the mean envelope calculation step of the method. The number of IMFs produced is not fixed and depends on the number of intrinsic modes of the data. This variant of AMP is referred to as empirical mode decomposition AMP (EMD-AMP).\\

\section{Empirical Evaluation}
\label{sec:eval}

\noindent \textcolor{black}{
One of the most common reasons that researchers extract features from time-series is to serve as inputs for machine learning algorithms. Therefore, to assess the performance of AMP, we chose the real-world application of time series clustering. First we perform an evaluation of the effectiveness of DFT-AMP, DWT-AMP, DWPT-AMP and EMD-AMP using synthetic data (in order that we have a ground to truth to assess against)}, showing that it outperforms traditional frequency domain and time domain based clustering techniques. We also show that across all variants, EMD-AMP is the most effective in partitioning data into groups with similar time-frequency patterns. 
With this demonstrated, EMD-AMP is then applied to a real world data set made up of the phone call logs of Massachusetts Institute of Technology (MIT) faculty and students \cite{eagle2009inferring}. The population of MIT individuals is clustered according to the IMFs they most express with scatter plots revealing two distinct groupings that correspond to different departments in which the staff and students work.

\subsection{Synthetic data}
\label{sec:synth_eval}

\noindent \textcolor{black}{Each of our experiments involves two distinct groups of labelled synthetic time-series. Each set includes a family of realizations generated by mixing two sinusoidal time-frequency patterns, with each set containing a distinct mix. Labels are removed, and the performance of all AMP variants is then evaluated by 1. extracting the time-frequency features obtained by each variant; 2. using these as inputs for cluster analysis; and 3. determining the extent to which the original set labels have been recovered. Clustering performance is} compared to using traditional time-frequency feature extraction methods and time domain based clustering approaches. The traditional methods considered for comparison are:

\noindent {\bf Fourier power clustering (Four. pow.).}  Here each time-series $\mathbf{x_{i}}$ is decomposed into a Fourier series using the DFT. The power of the components at each frequency is used as a feature for clustering. For consistency with the DFT-AMP approach, 1022 Fourier components are used. \\

\noindent {\bf Wavelet coefficient clustering (wav. coef.).} 
Each time-series is decomposed via the DWT using the Haar wavelet and the coefficient values for each wavelet term are used as features for clustering. Again, each time-series is decomposed into 1022 basis vectors. 

\noindent {\bf Euclidean distance clustering (Euc.).}
Because of its simplicity and the fact such an approach can give excellent results \cite{keogh2003need}, we consider a clustering approach based on the Euclidean distances between time-series.  

\noindent {\bf Dynamic time warping distance clustering (DTW)}
\textcolor{black}{
As evidence suggests that DTW distance is the most effective distance measure for classification tasks \cite{ding2008querying} we investigate the performance of DTW distance between time-series on the clustering experiments using the standard DTW algorithm \cite{berndt1994using}.}

Four. pow., wav. coef., Euc. and DTW were used to obtain the results in Figure \ref{fig:intermittency_examples} in the introduction.

\subsubsection{Data set generation}
\label{sec:syn_generation}
\noindent
\textcolor{black}{
To produce our synthetic dataset we use a stochastic data generation model that underpins a model for the times at which emails are sent \cite{malmgren2008poissonian}.}
Such a model allows us to control the the intermittency as well as the stationarity of the data, which is particularly useful given real world human activity data is often non-stationary \cite{malmgren2009universality, stehle2010dynamical, zhao2011social}. We \textcolor{black}{generate three experimental }datasets, the first in which all data is stationary ({\bf Syn1}), the second in which non-stationary data is considered ({\bf Syn2}), and the final set where noise is added to non-stationary data ({\bf Syn3}). In order to investigate the impact of intermittency on the results we also vary the amount of intermittency the data exhibits. 

To generate data sets of intermittent time-series with known time-frequency features (and hence known cluster memberships) we, in the first instance, generate temporal point process data \cite{daley2007introduction} with a prescribed generating function which controls the time-frequency patterns in the data. Synthetic time-series are then created by mapping the point process data to a continuous function by convolving the data with a kernel \cite{silverman1986density} as follows

 \begin{equation}
x_{i}(t)=\frac{1}{\theta_{i}} \sum\limits_{k=1}^{n_{i}} K \left(\frac{t-t_{ik}}{h} \right),
\label{eq:kd_ind}
\end{equation}

\noindent where, $t$ is time, $t_{ik}$ is the k$^{\mbox{th}}$ point process event attributed to time-series $i$, $\theta_{i}$ the total number of events generated and $K$ is the standard normal density function with bandwidth $h$. Function (\ref{eq:kd_ind}) is then sampled at $n$ equally spaced points in time to obtain the discrete time-series $\mathbf{x_{i}}$. The $t_{ik}$'s are generated using a non-homogeneous Poisson process \cite{ross2006introduction}. By utilising the rate function for the non-homogeneous Poisson process we can prescribe different time-frequency patterns in the synthetic time-series data. 
\textcolor{black}{
In each dataset, two equally sized groups of time-series are considered. Both express the same two time-frequency patterns but in different degrees. The data for each group is generated using the following rate functions which are sums of two almost periodic functions:}
\begin{eqnarray}
\text{\textbf{group 1:}}\: \nonumber \lambda_{1}(t)&=\varphi \left( \gamma  \text{sin}^{2}(\frac{\pi t}{T_{1}(t)}) + (1-\gamma) \text{sin}^{2}(\frac{\pi t}{T_{2}(t)}) \right ) \: \\
 \text{if} \: i\le m/2, \label{eq:population1} \\
\text{\textbf{group 2:}}\: \nonumber \lambda_{2}(t)&=\varphi \left( (1-\gamma) \text{sin}^{2}(\frac{\pi t}{T_{1}(t)}) + \gamma  \text{sin}^{2}(\frac{\pi t}{T_{2}(t)}) \right) \: \\
\text{if} \: i> m/2,
\label{eq:population2}
\end{eqnarray}

\textcolor{black}{
The amplitude coefficient, $\varphi$, effectively controls how \textit{intermittent} the time-series is (smaller values lead to more intermittent time-series). While the `mixing' parameter, $\gamma \in [0,0.5]$, allows us to control how similar the two groups of realizations are. If  $\gamma = 0.5$ both sets will be expressing the same mix of time-frequency patterns and won't be able to be distinguished. Otherwise the groups will express the same time-frequency patterns to different degrees, and as $\gamma$ approaches zero will become increasingly distinct. The periods $T_{1}$ and $T_{2}$ of each sinusoidal function are defined as:}
\begin{eqnarray}
T_{1}(t)&=T_{1}'+\alpha_{1} t, \: \: \: T_{2}(t)&=T_{2}'+\alpha_{2} t, \label{eq:period_func} 
\label{eq:periods}
\end{eqnarray}
where $T_{1}'$ and $T_{2}'$ are constants. The coefficients $\alpha_{1}$ and $\alpha_{2}$ act to allow non-stationary scenarios to be considered where the period of oscillation of the rate functions (\ref{eq:population1}) and (\ref{eq:population2}) change with time. 

\textcolor{black}{
These parameters allow us to produce three distinct experimental datasets, \textbf{Syn1, Syn2} and \textbf{Syn 3}, against which we can evaluate performance. The datasets as described below:}
\label{sec:dgen}
\begin{description}
\item[{\bf Syn1:}] Stationary, No Noise\\
In this dataset, $\alpha_{1}$ and $\alpha_{2}$ from equation (\ref{eq:periods}) are both set to 0 to ensure that the period of the rate functions of the non-homogeneous Poisson processes is fixed. $T'_{1}$ and $T'_{2}$ are set to 2 and 8 respectively.
\item[{\bf Syn2:}] Non-Stationary, No Noise\\
As for {\bf Syn1}, except $\alpha_{1}= 0.0078$ and $\alpha_{2}=  0.0314$ which ensures that the period of the rate functions are an increasing linear function of time. In particular, the period of rate function $\lambda_1(t)$ (\emph{resp.} $\lambda_2(t)$) from equations (\ref{eq:population1}) and (\ref{eq:population2}) ranges from 2 (4) at $t=0$ to 4 (8) at $t=255$. This means that the time-series are characterised by time-frequency patterns with period which increases with time. 
\item[{\bf Syn3:}] Non-Stationary, Noisy\\
As for {\bf Syn2}, except noise is incorporated in one tenth of the time-series within the set. In particular, $m/10$ time-series were selected at random. Of the temporal events from which these time-series were formed, 50 are selected at random and an additional 41 events (equally distributed over a period of 0.02 time units) are introduced starting from the selected time point. This gives a total of 2050 additional events per time-series selected. These manifests themselves as `spikes' in the time-series where the value of the dependent variable rises and falls extremely quickly.

\end{description}

\subsubsection{Results}
\label{sec:syn_results}
\begin{figure}
 \centering
 \hspace{0.05in} \textbf{Syn. 1} \hspace{0.52in} \textbf{Syn. 2}  \hspace{0.51in} \textbf{Syn. 3} \\
 Performance against `mixing parameter' $\gamma$\\
 \hspace{0.06in}\raisebox{0.1in}{\includegraphics[width=0.015\textwidth]{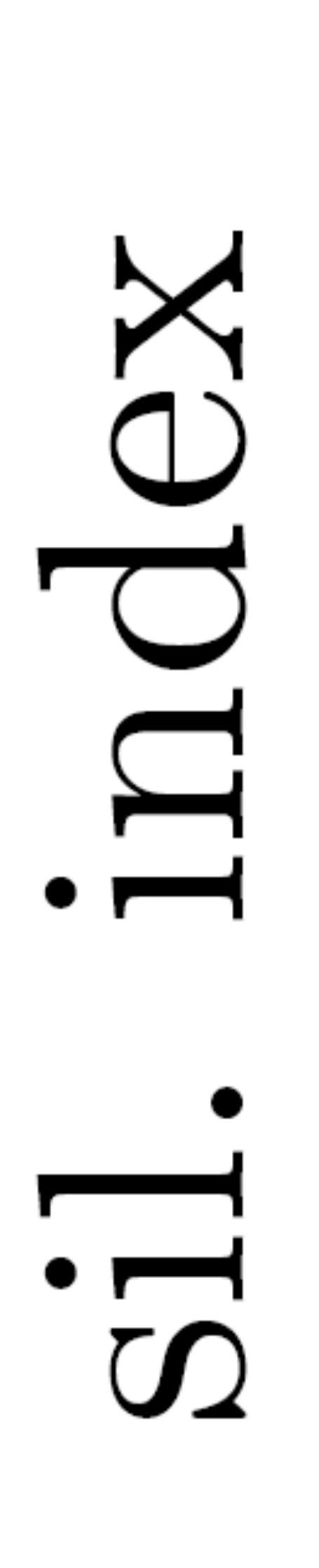}}\hspace{-0.06in}
 \includegraphics[width=0.14\textwidth]{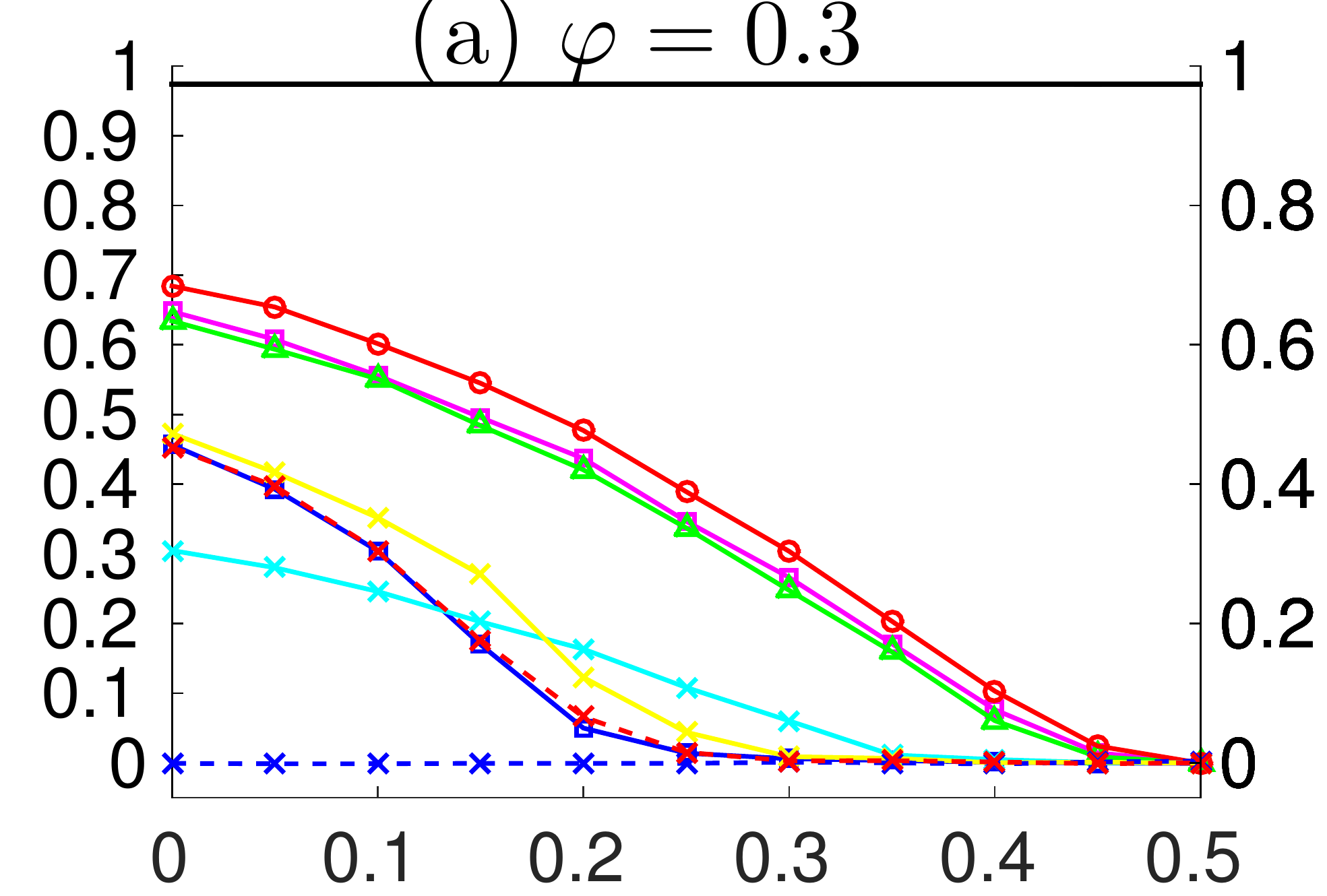}
 \includegraphics[width=0.14\textwidth]{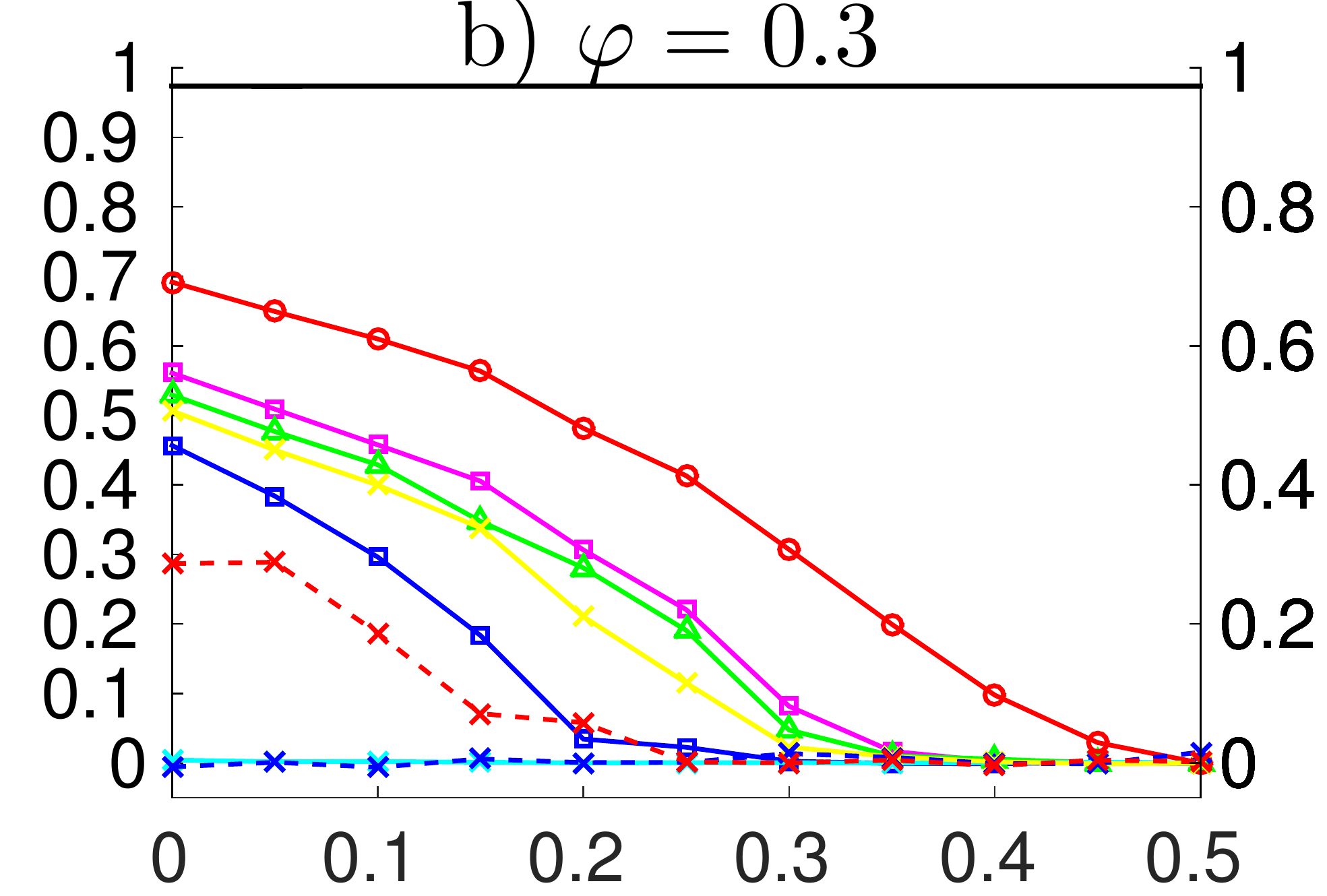}
 \includegraphics[width=0.14\textwidth]{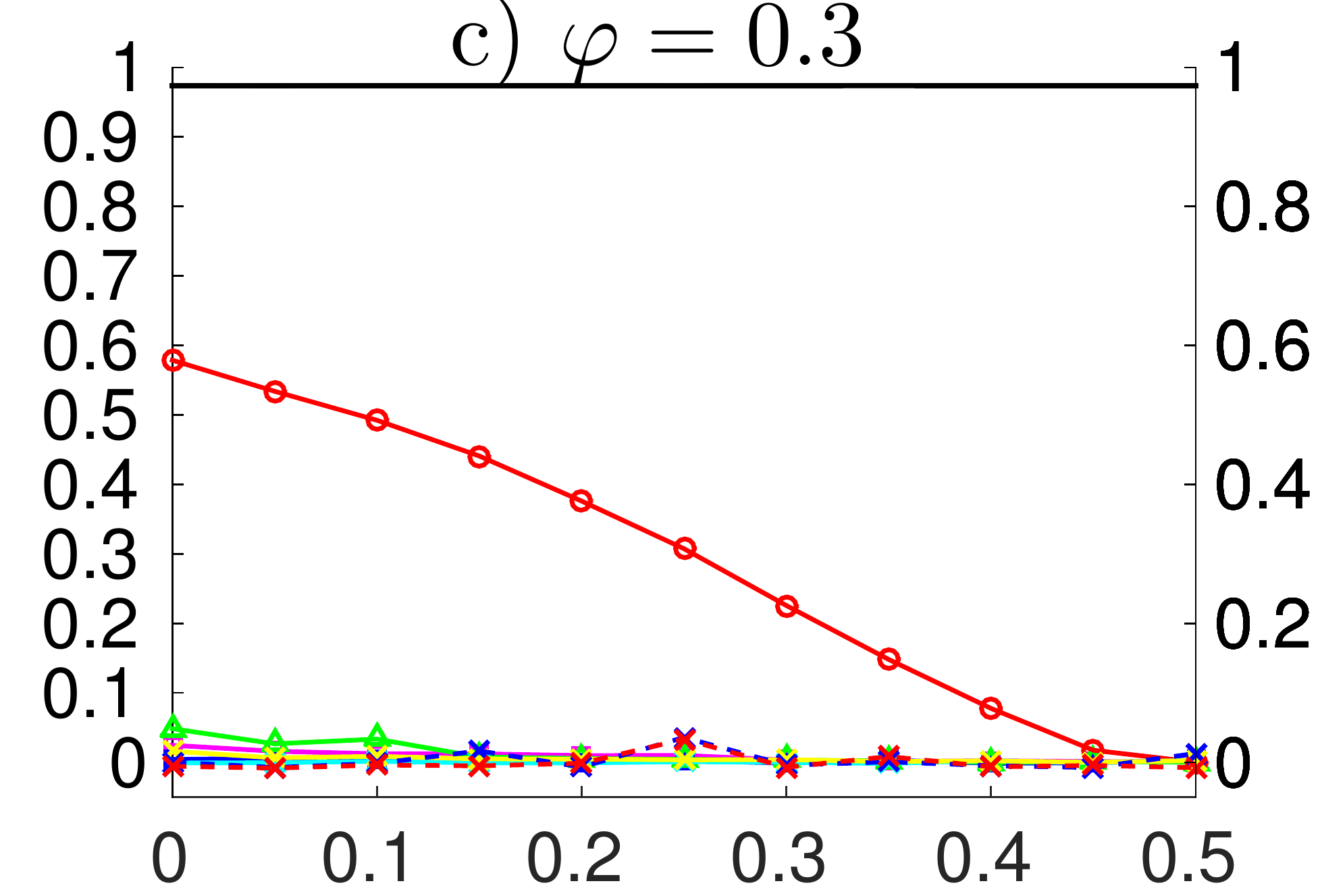}
 \hspace{-0.04in}\raisebox{0.28in}{\includegraphics[width=0.008\textwidth]{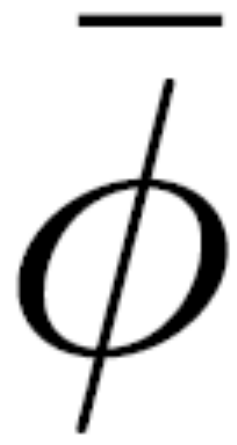}}\hspace{0.04in}\\
 \raisebox{0.1in}{\includegraphics[width=0.01\textwidth]{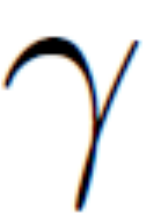}}\\
 Performance against \textcolor{black}{`intermittency'} parameter $\varphi$\\
 \hspace{0.06in}\raisebox{0.1in}{\includegraphics[width=0.015\textwidth]{sil_index.pdf}}\hspace{-0.06in}
 \includegraphics[width=0.14\textwidth]{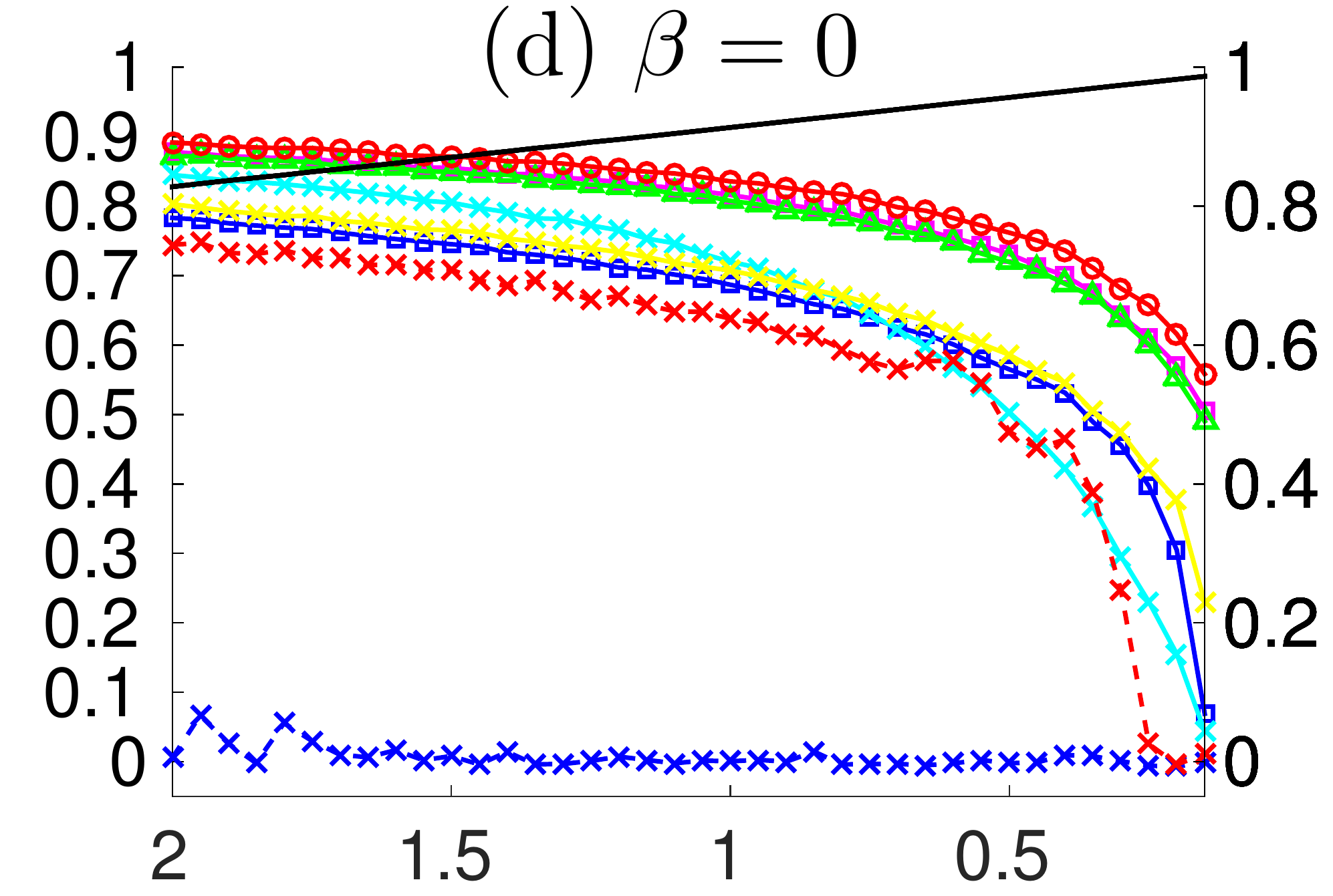}
 \includegraphics[width=0.14\textwidth]{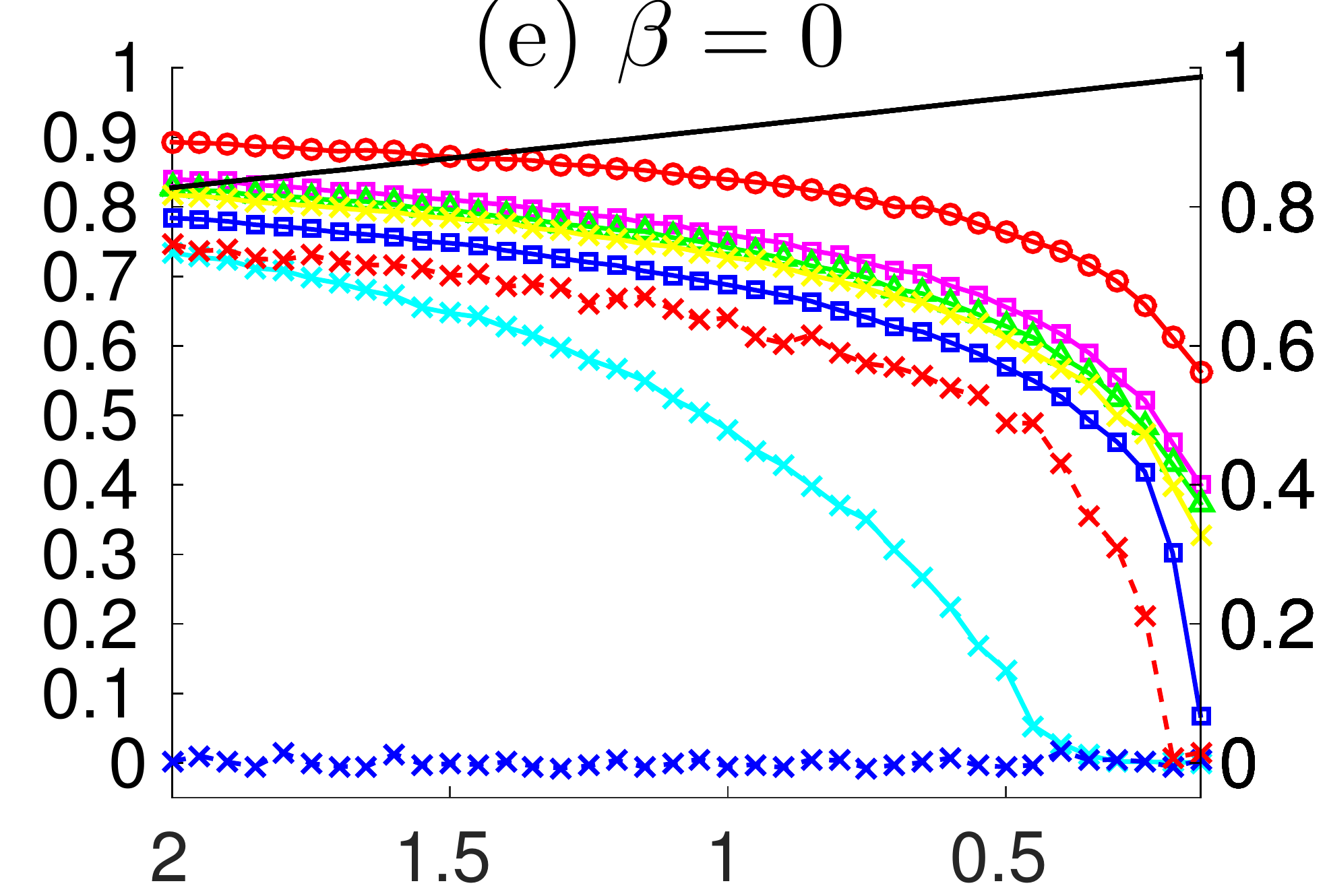}
 \includegraphics[width=0.14\textwidth]{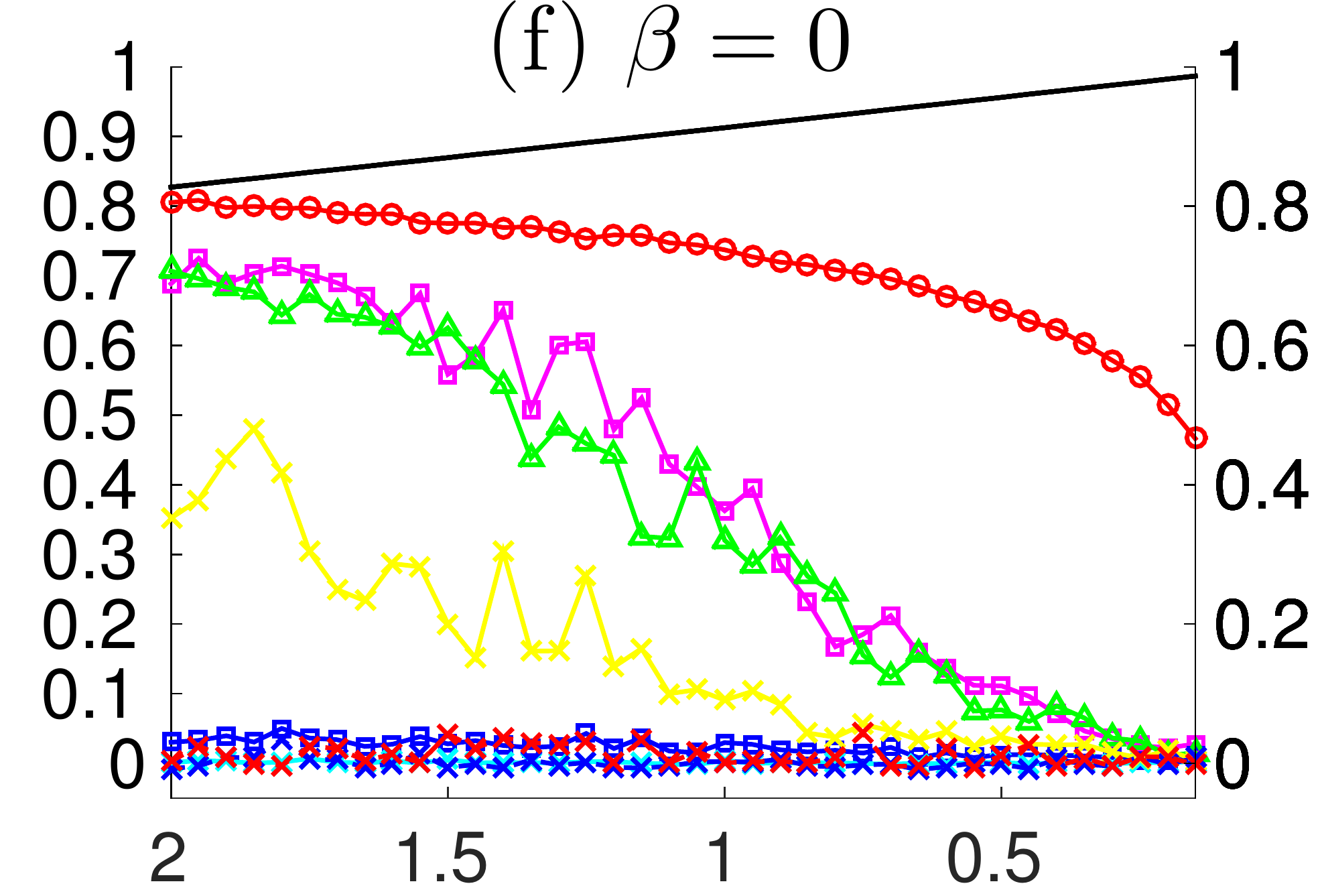}
 \hspace{-0.04in}\raisebox{0.28in}{\includegraphics[width=0.008\textwidth]{phi.pdf}}\hspace{0.04in}\\
 \raisebox{0.1in}{\includegraphics[width=0.01\textwidth]{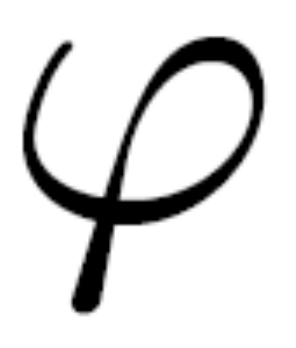}}\\
 \vspace{0.1in}
 Performance against `mixing' parameter $\gamma$\\
 \hspace{0.06in}\raisebox{0.07in}{\includegraphics[width=0.015\textwidth]{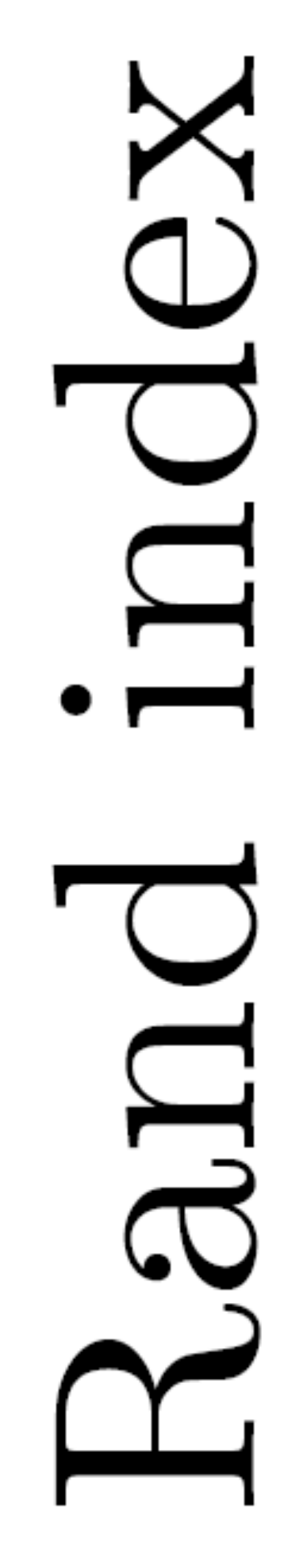}}\hspace{-0.06in}
 \includegraphics[width=0.14\textwidth]{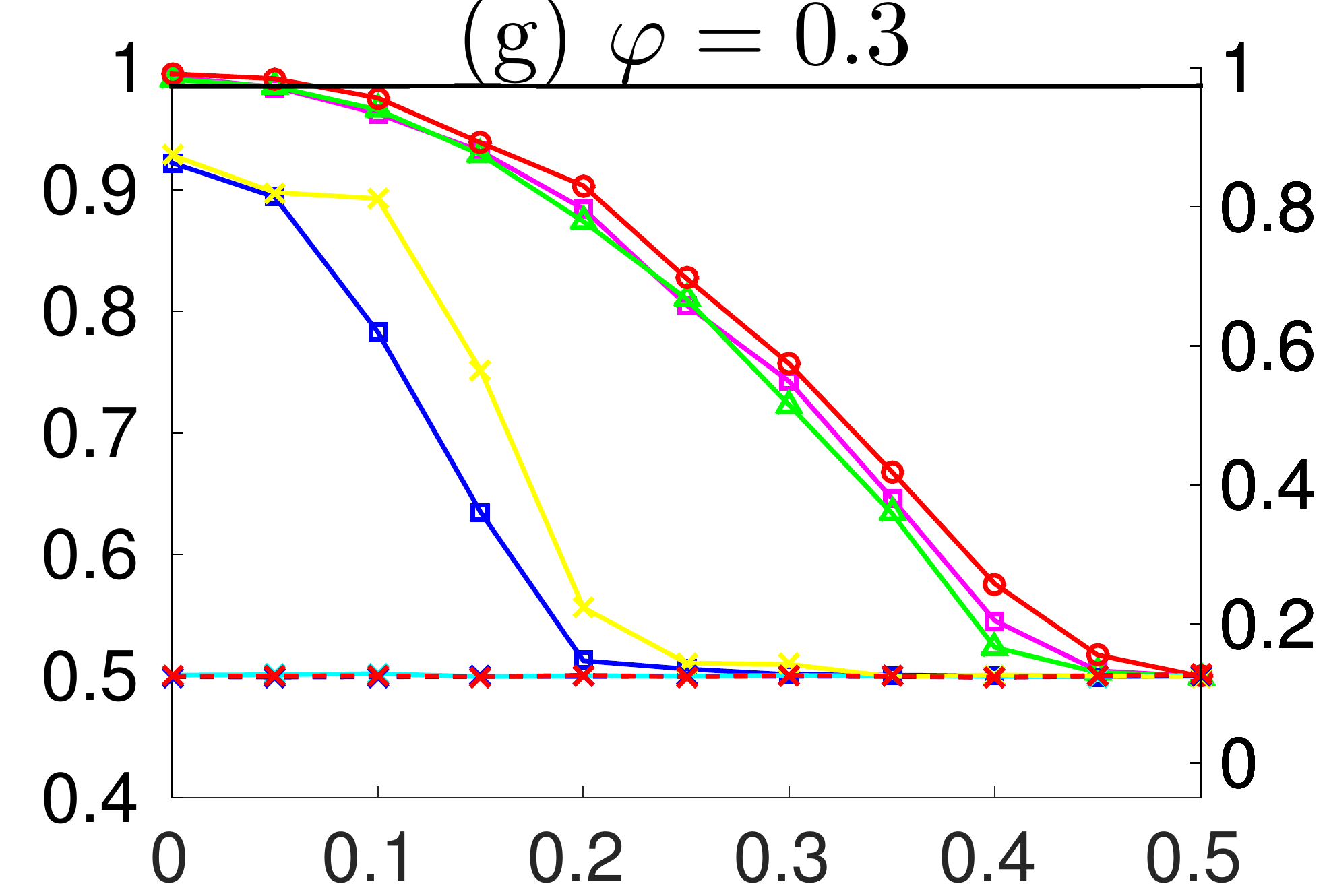}
 \includegraphics[width=0.14\textwidth]{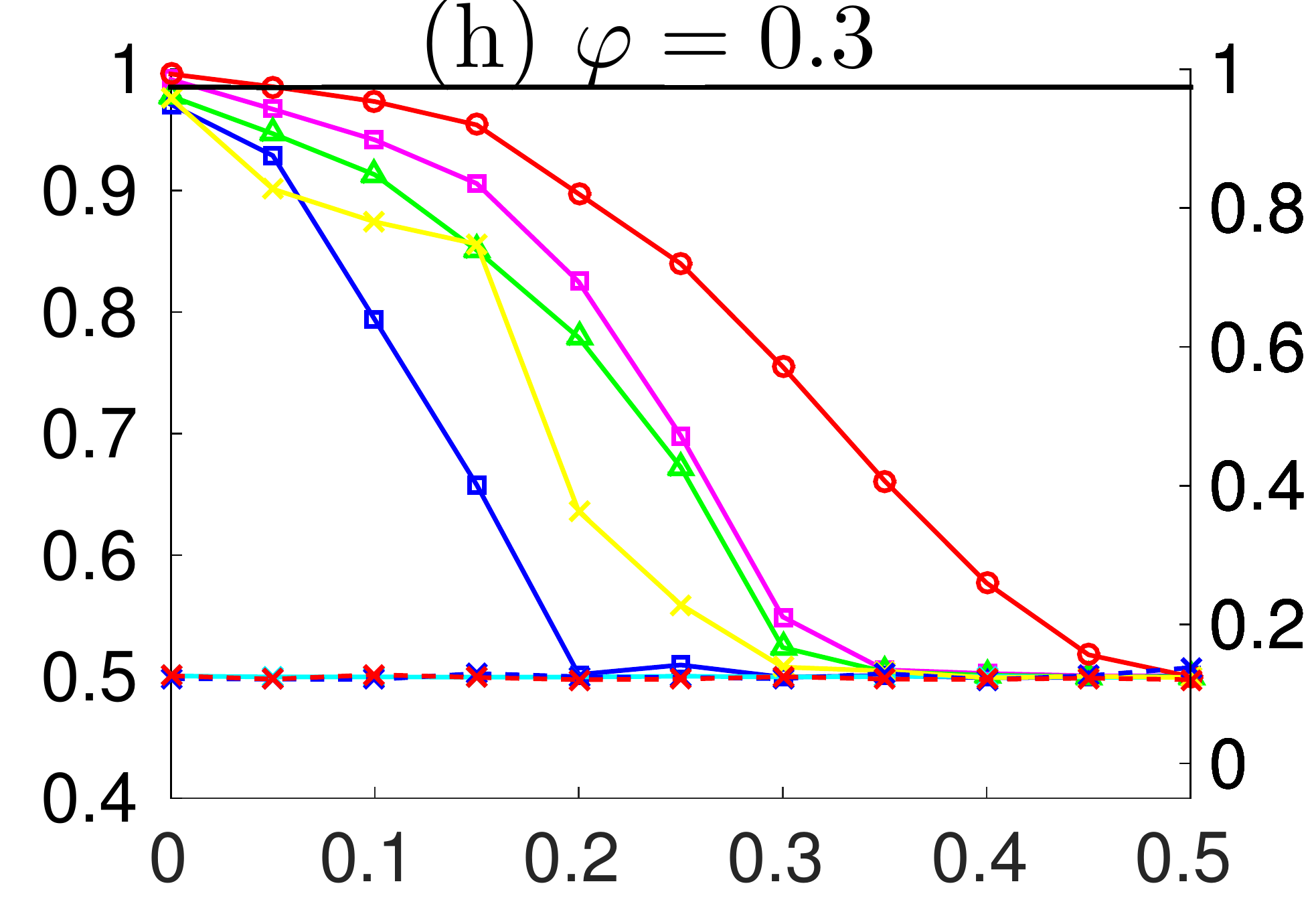}
 \includegraphics[width=0.14\textwidth]{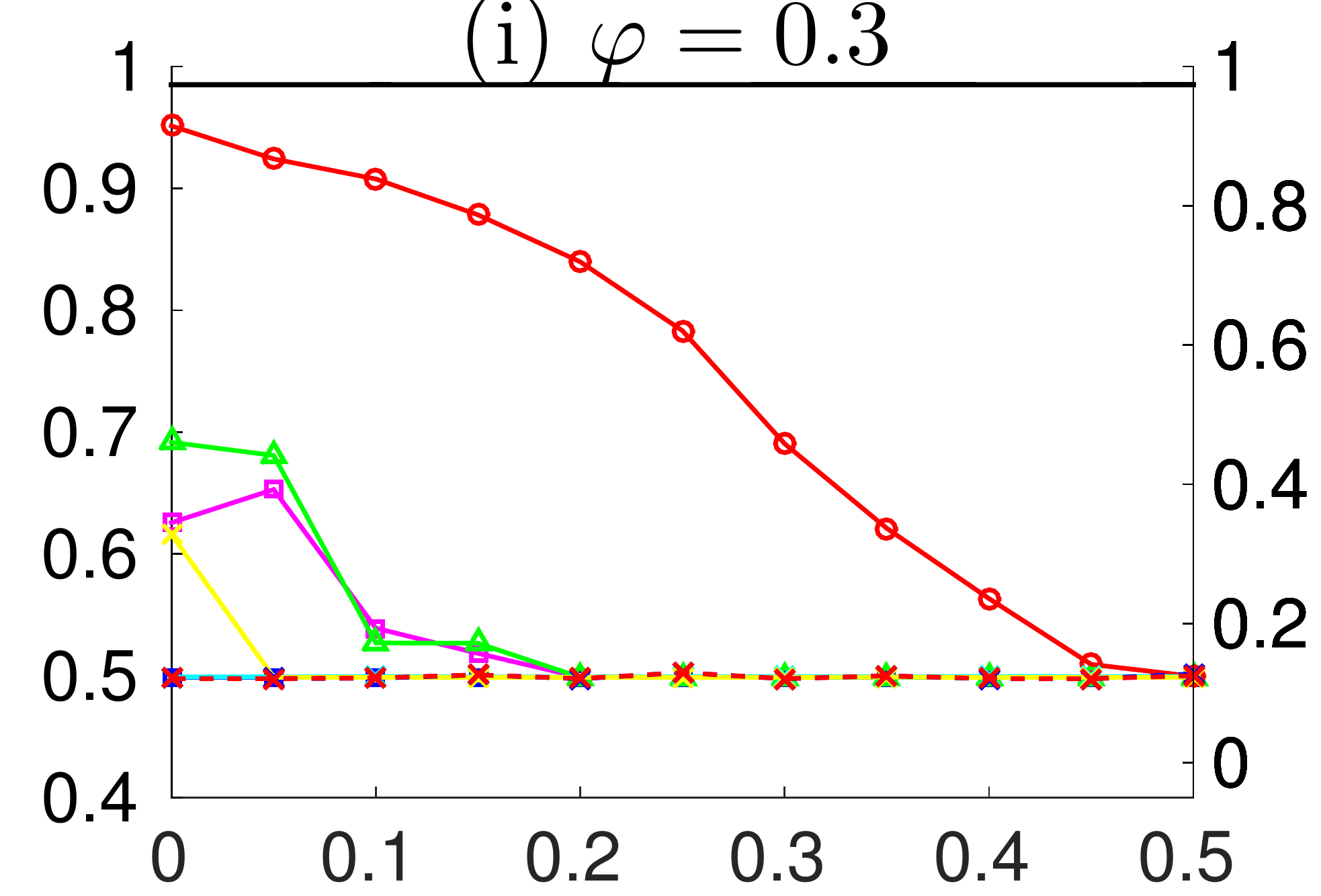}
 \hspace{-0.04in}\raisebox{0.28in}{\includegraphics[width=0.008\textwidth]{phi.pdf}}\hspace{0.04in}\\
 \raisebox{0.1in}{\includegraphics[width=0.01\textwidth]{gamma.pdf}}\\
 Performance against \textcolor{black}{`intermittency'} parameter $\varphi$\\
 \hspace{0.06in}\raisebox{0.07in}{\includegraphics[width=0.015\textwidth]{rand_index.pdf}}\hspace{-0.06in}
 \includegraphics[width=0.14\textwidth]{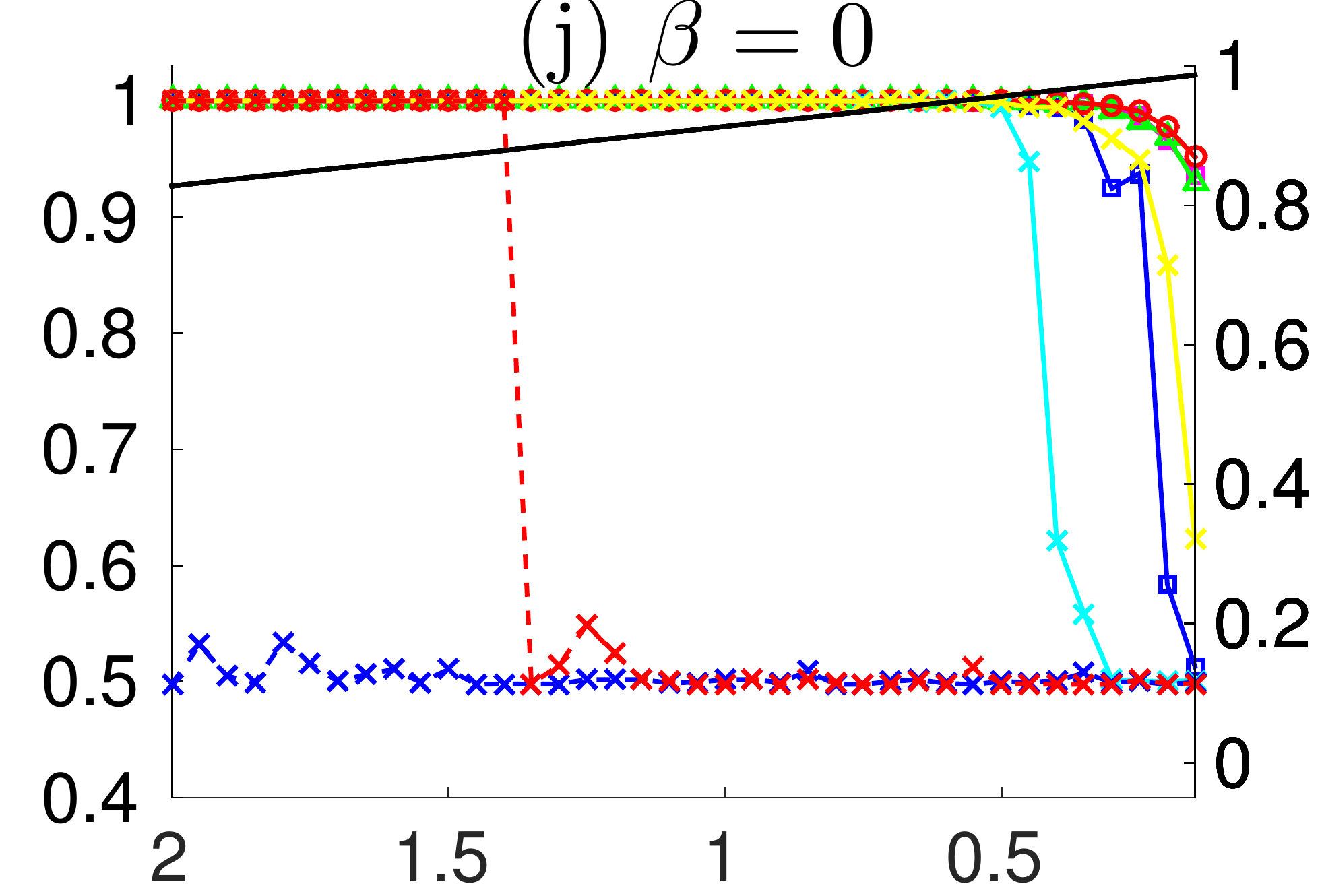}
 \includegraphics[width=0.14\textwidth]{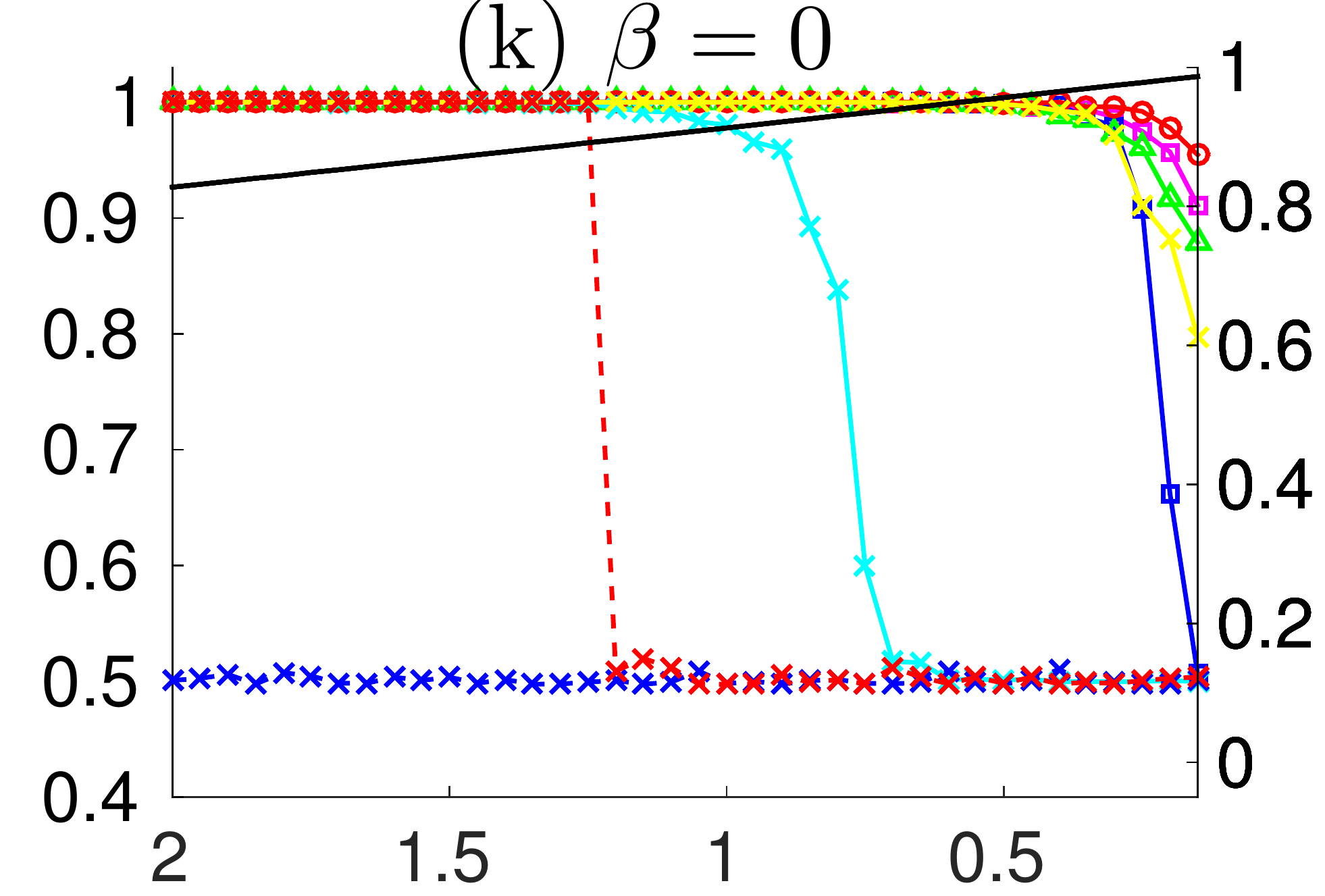}
 \includegraphics[width=0.14\textwidth]{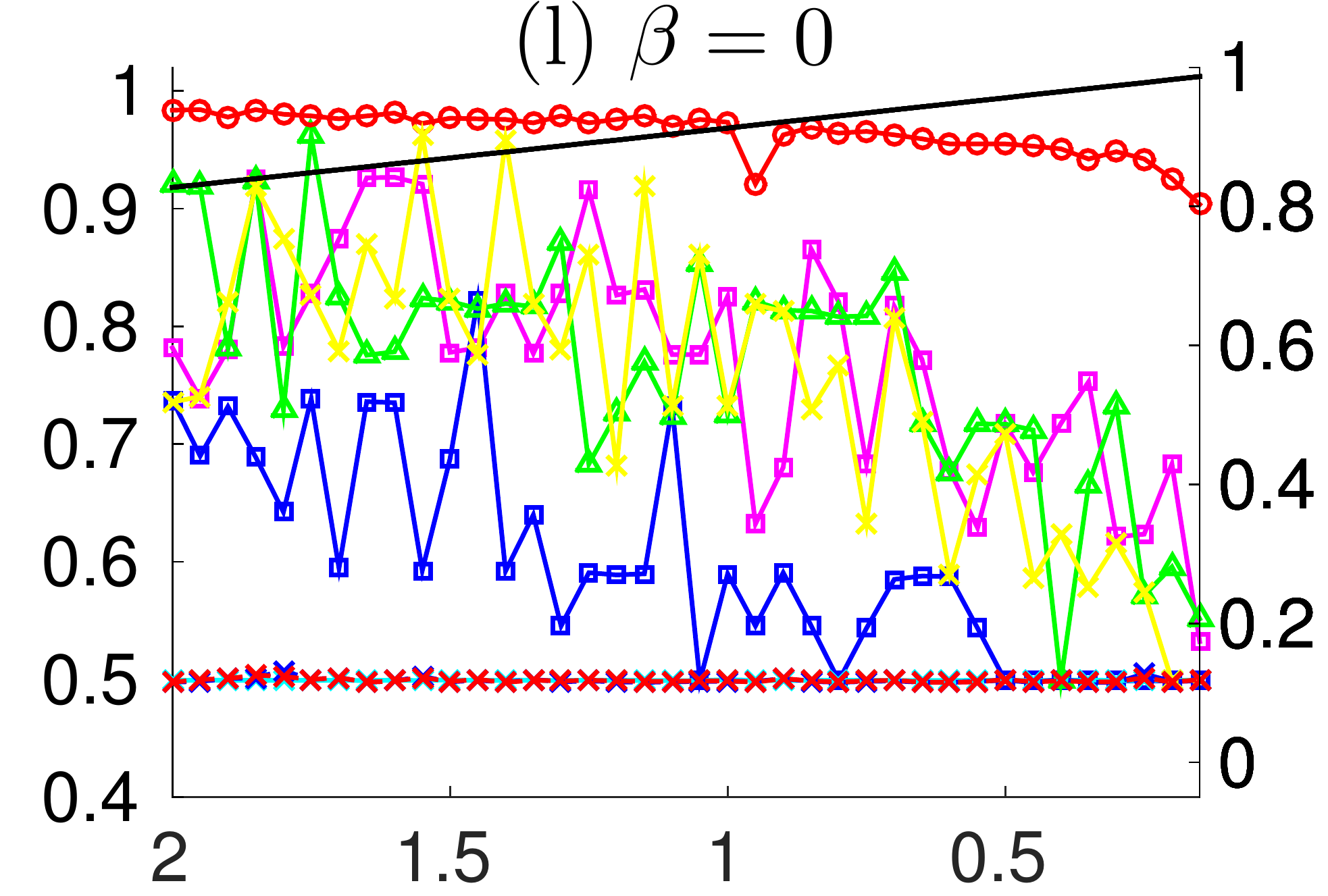}
 \hspace{-0.04in}\raisebox{0.28in}{\includegraphics[width=0.008\textwidth]{phi.pdf}}\hspace{0.04in}\\
 \raisebox{0.1in}{\includegraphics[width=0.01\textwidth]{varphi.pdf}}\\
 \includegraphics[width=0.2\textwidth]{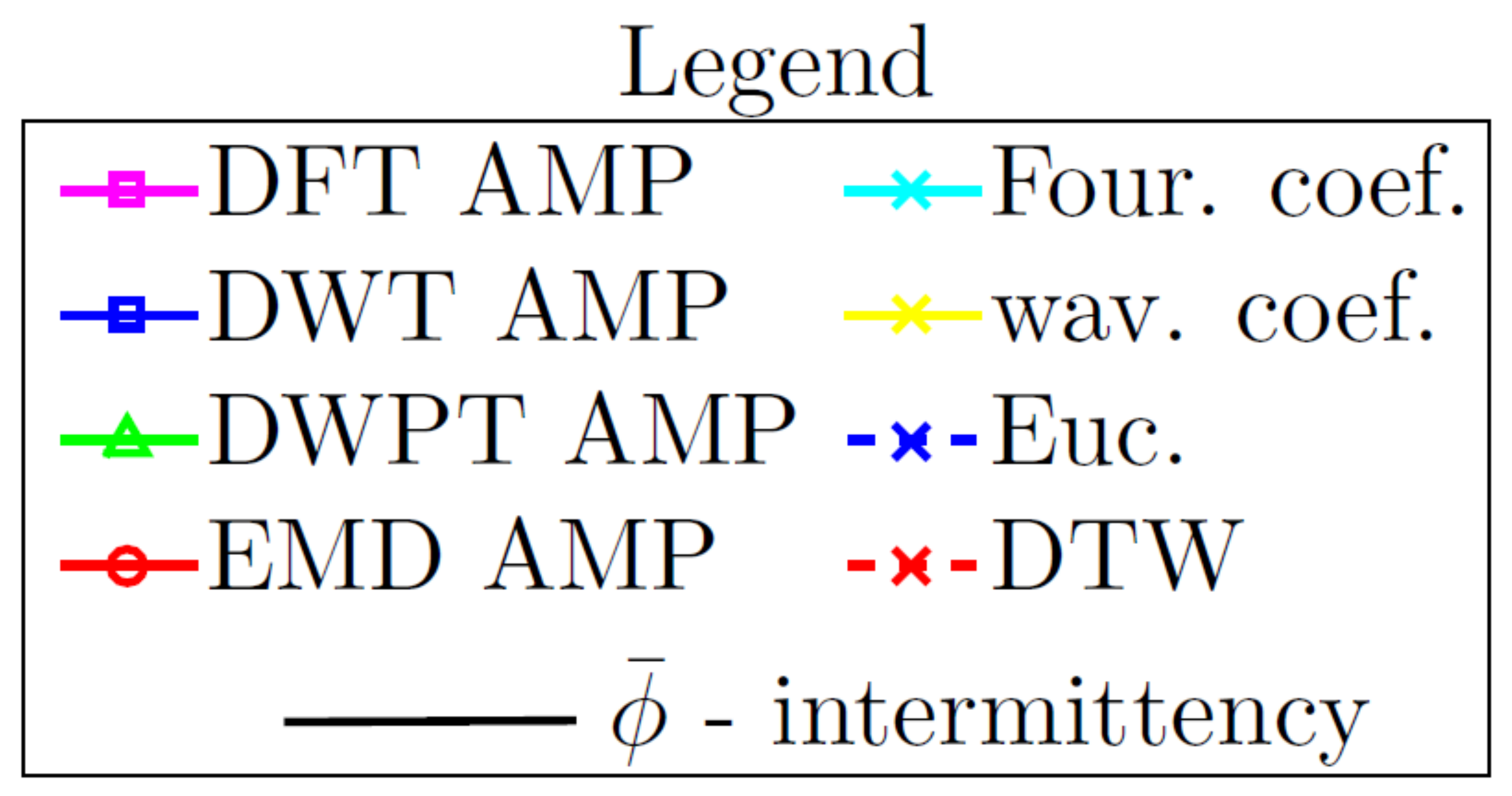}
  
  \caption{\small Plots (a)-(f) show the mean silhouette index score for the two dimensional representations of the feature values obtained using all methods detailed in Sections \ref{sec:method} and \ref{sec:eval}. Scores are plotted as a function of the mixing parameter $\gamma$ (a-c) and amplitude parameter $\varphi$ (d-f). \textcolor{black}{Note that intermittence \textit{increases} as $\varphi$ decreases.}  Plots (g)-(l) show the Rand index scores obtained by comparing the results of $k$-means clustering applied to the full set of coefficient scores for each method to the true clustering. The average intermittence score $\bar{\phi}$ (equation (\ref{eq:interittence})) for all time-series in a set is also shown in each plot (value given in the right axes). For each value of $\gamma$ and $\varphi$ considered, ten simulations were run and an average taken for the mean silhouette score, Rand index and mean intermittence score. Parameter values for all simulations are $m=4000$, $h=0.05$ and where given in Section \ref{sec:syn_generation} and in the plots. All simulations were run for 256 time units.}
 \label{fig:beta_varphi}
 \end{figure}

\noindent \textcolor{black}{Results of the} synthetic experiments are shown in Figure \ref{fig:beta_varphi}. The performance of the methods presented in this paper are measured firstly by the mean silhouette score \cite{rousseeuw1987silhouettes} for all data points in a set against the true clustering in two dimensions (obtained via classical MDS where applicable). A score of 1 indicates maximal distance between the two true clusters (i.e. between the data points of groups 1 and 2), with 0 corresponding to maximal mixing between the clusters. The  performance is also measured via the Rand index \cite{rand1971objective} between the true clustering and that obtained from the application of $k$-means ($k=2$) \cite{macqueen1967some} to the full set of features outputted by each method. Here, 1 corresponds to perfect agreement between the $k$-means results and the true clustering. For each data set the effects of varying the mixing parameter $\gamma$ and amplitude parameter $\varphi$ (equations (\ref{eq:population1}) and (\ref{eq:population2})) are also considered.

The results illustrate that all AMP variants consistently outperform state of the art techniques (wav. coef., Four. pow, Euc. and DTW) in every plot except for the DWT-AMP variant, the performance of which is comparable to wav. coef. and Four. pow. Of all variants, EMD-AMP is the best performer. This is particularly noticeable when the data sets are both non-stationary and contain noise (see plots (c,f,i,l). All frequency domain based methods outperform the time domain based methods (Euc. and DTW). \textcolor{black}{This latter results highlights that DTW and Euclidean distance based methods are not suitable for the clustering of intermittent data}. As expected, the performance of every method decreases as the mixing parameter $\gamma$ increases and amplitude parameter $\varphi$ (which is inversely related to intermittence) decreases.

\textcolor{black}{To aid our explanation as to why AMP appears to be performing so well }in producing accurate time-frequency features for intermittent data and why, of all the variants, EMD-AMP produces the most accurate features, we consider a specific instance of \textbf{Syn3}, the non-stationary and noisy data set. We use the same parameter values for synthetic data generation as those used in Figure \ref{fig:beta_varphi}l \textcolor{black}{and set $\gamma=0$ (ensuring that the time-frequency patterns of both groups are as distinct as can be, with each group expressing completely different time-frequency patterns) }. The period of oscillation of group 1 time-series ranges linearly from 2 at $t=0$ to 4 at $t=255$ and group 2 data from 4 to 8. 

\begin{figure}[htbp]
 \centering
 Individual time-series examples\\
  \hspace*{-0.7cm}
  \includegraphics[width=0.53\textwidth,]{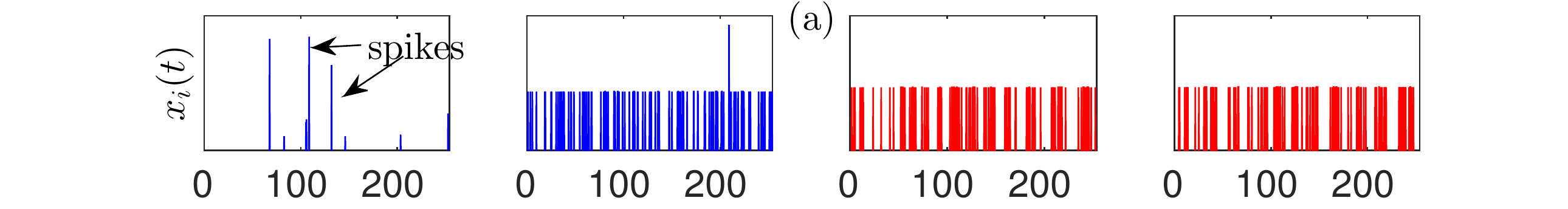}\\
  \vspace{0.2in}
  \includegraphics[width=0.23\textwidth,]{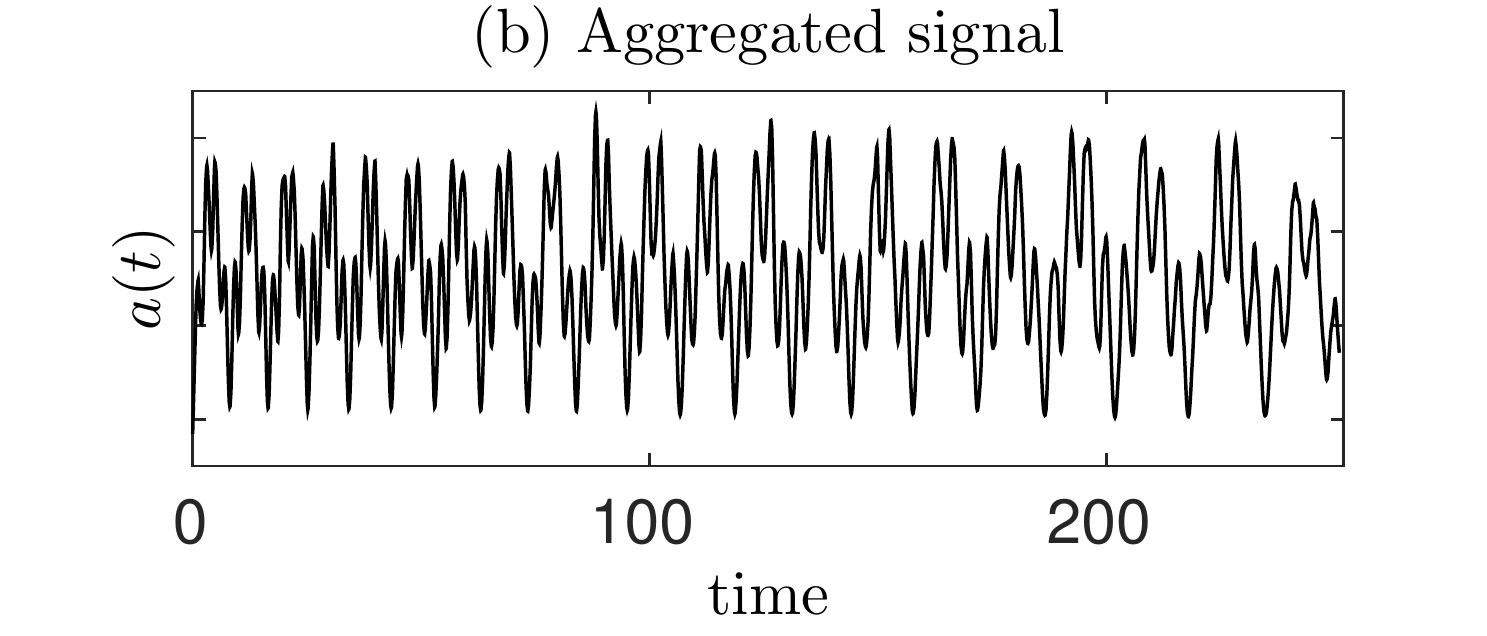}
  \includegraphics[width=0.23\textwidth,]{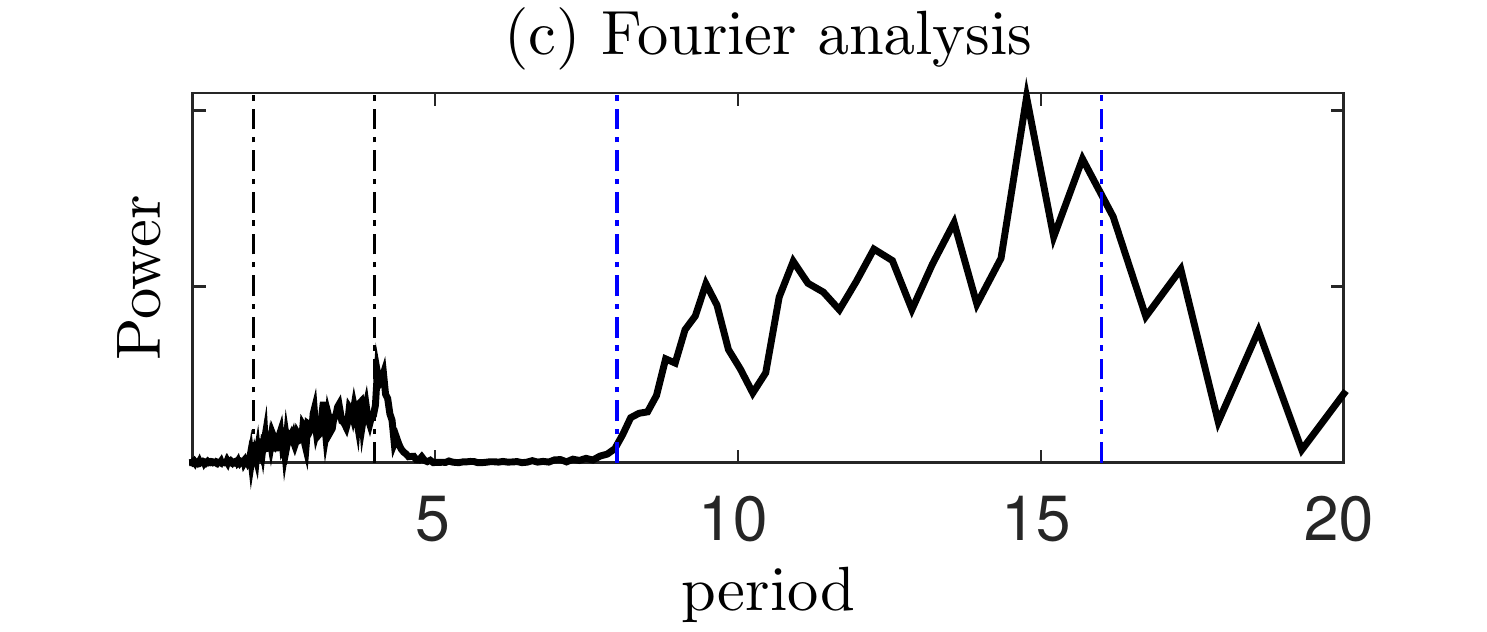}
  \hspace{0.5in}
  \includegraphics[width=0.23\textwidth,]{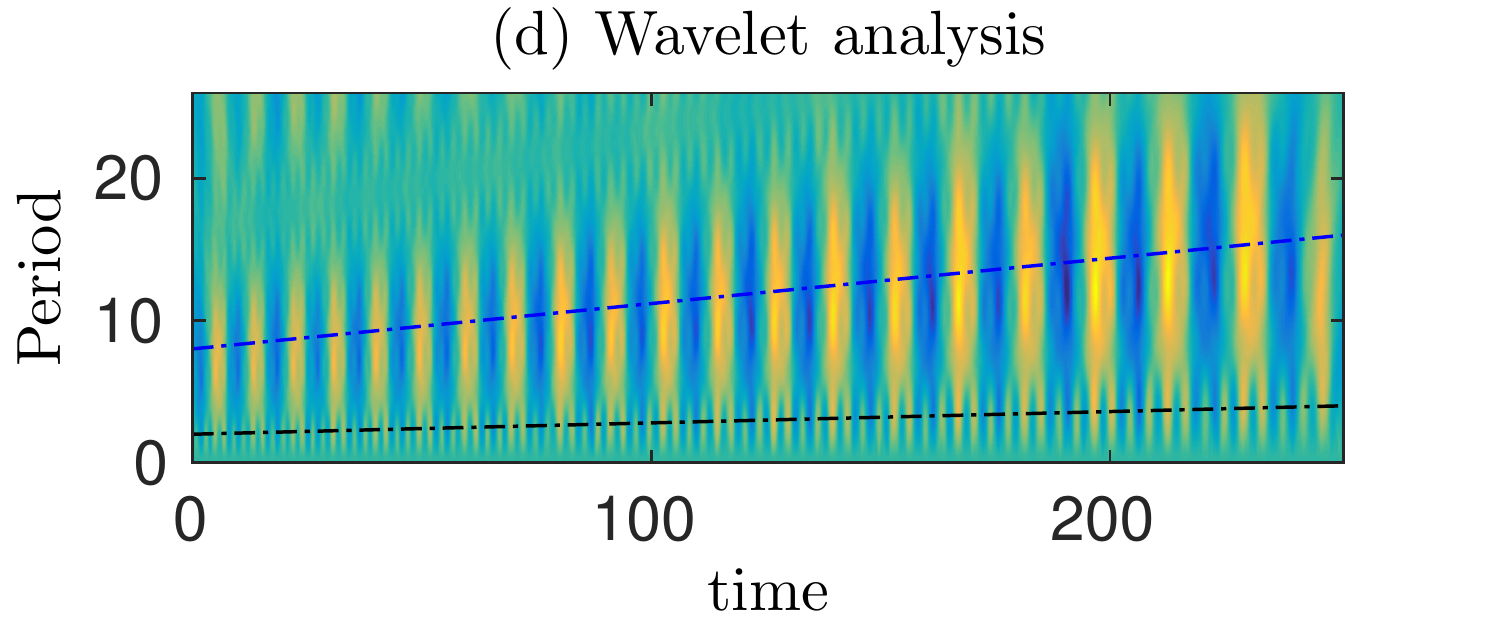}
  \includegraphics[width=0.23\textwidth,]{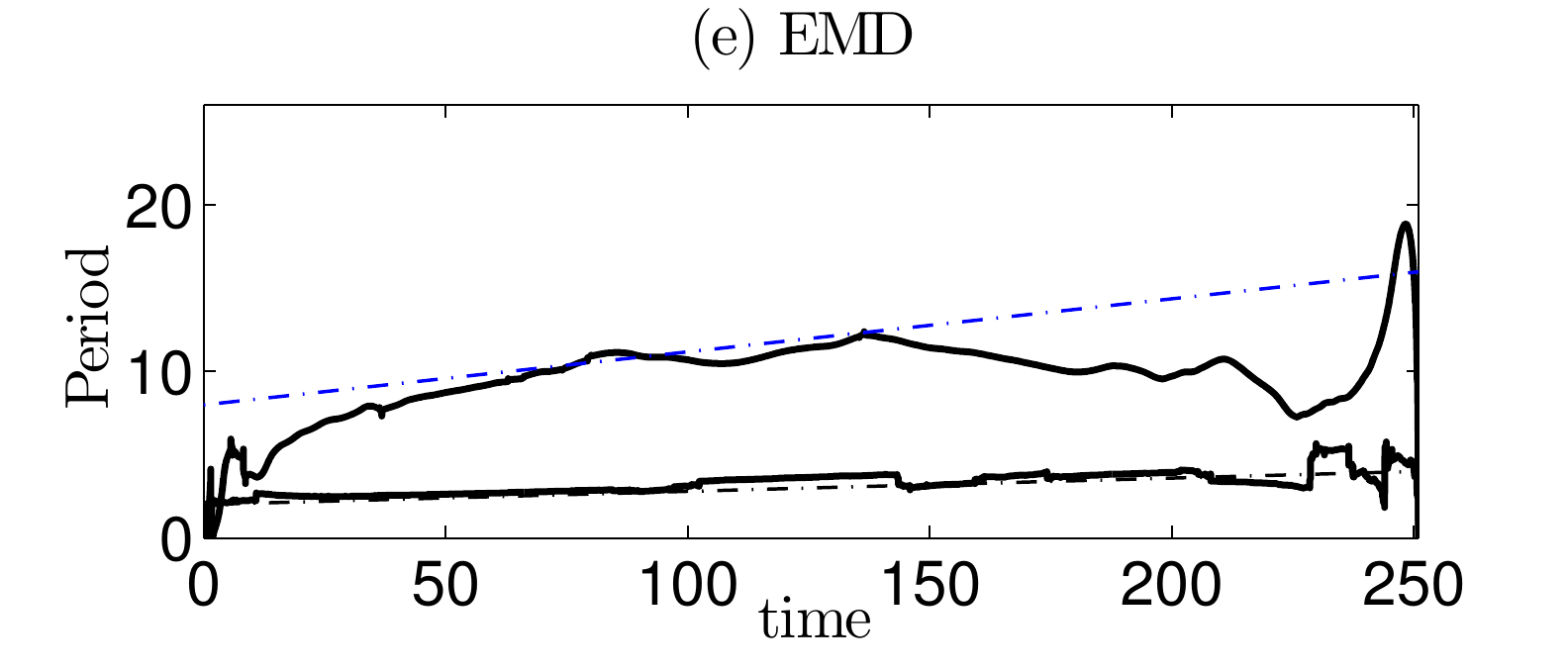}
    \caption{\small The top plots shows four individual time-series, two from group 1 (blue lines) and two from group 2 (red lines) illustrating the intermittent nature of the data. The first of these was generated from data containing noise which manifests itself as `spikes' (some examples marked) in the time-series. The plot also shows the aggregate (equation (\ref{eq:aggregate_time})) obtained by combining all 4000 intermittent time-series in the set. The non-intermittent aggregate permits wavelet and empirical model decompositions which reveal the two underlying time-frequency patterns (indicated by blue and red broken lines in the spectrum) of the data set. Note, the edge effects in the EMD plot are artefacts resulting from the discrete Hilbert transform of the IMFs. The Fourier spectra is also shown and this picks up the range of the frequencies of the two time-frequency patterns (indicated by blue and red broken lines).}
 \label{fig:nonstationary_noise_agg_signal_plus_analysis}
 \end{figure}

A sample of individual time-series from each group, together with the aggregate and spectrograms obtained from its decomposition are shown in Figure \ref{fig:nonstationary_noise_agg_signal_plus_analysis}. Despite the intermittent nature of the individual time-series, the aggregate is clearly non-intermittent (Figure \ref{fig:nonstationary_noise_agg_signal_plus_analysis}b). Furthermore, the noise present in some time-series has been suppressed by the aggregation process. The non-intermittence of the aggregate time-series is one of the strengths of the AMP approach as it permits decompositions which do not contain spurious signals corresponding to ringing artefacts. In particular wavelet and EMD decompositions reveal only the two time-frequency patterns (one with period ranging from 2 to 4 and the other with period from 4 to 8) which are present within the data set (Figures \ref{fig:nonstationary_noise_agg_signal_plus_analysis}d and \ref{fig:nonstationary_noise_agg_signal_plus_analysis}e respectively). Because the aggregated signal is non-stationary with time varying frequency components, the Fourier spectrum picks up the range in frequencies of the underlying time-frequency patterns \textcolor{black}{but gives no indication as to \textit{when} different frequencies are present in the aggregate (Figure \ref{fig:nonstationary_noise_agg_signal_plus_analysis}c)}. 

\begin{figure}[htbp]
 \centering
  \includegraphics[width=0.45\textwidth,]{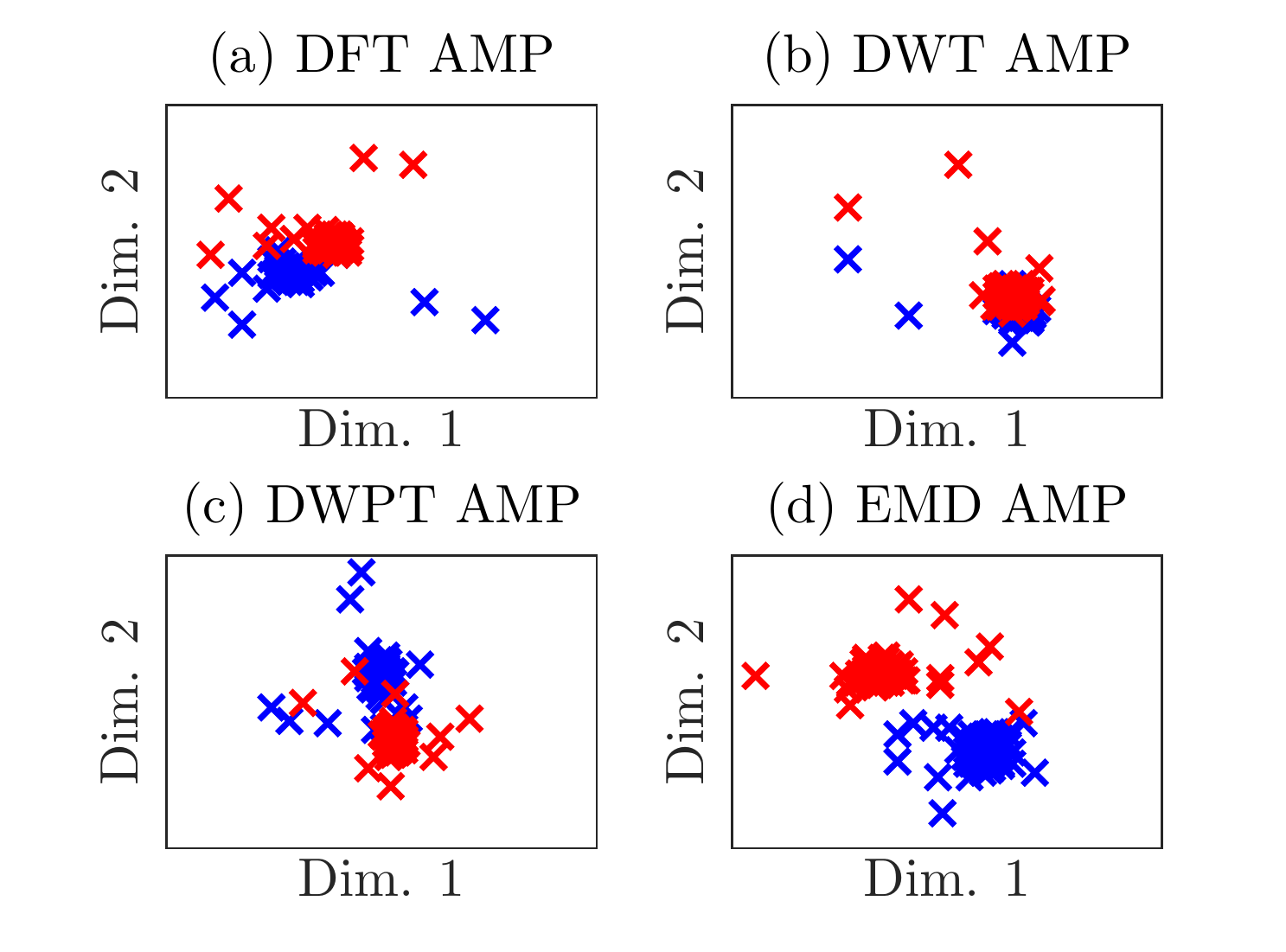}\\
  \includegraphics[width=0.1\textwidth,]{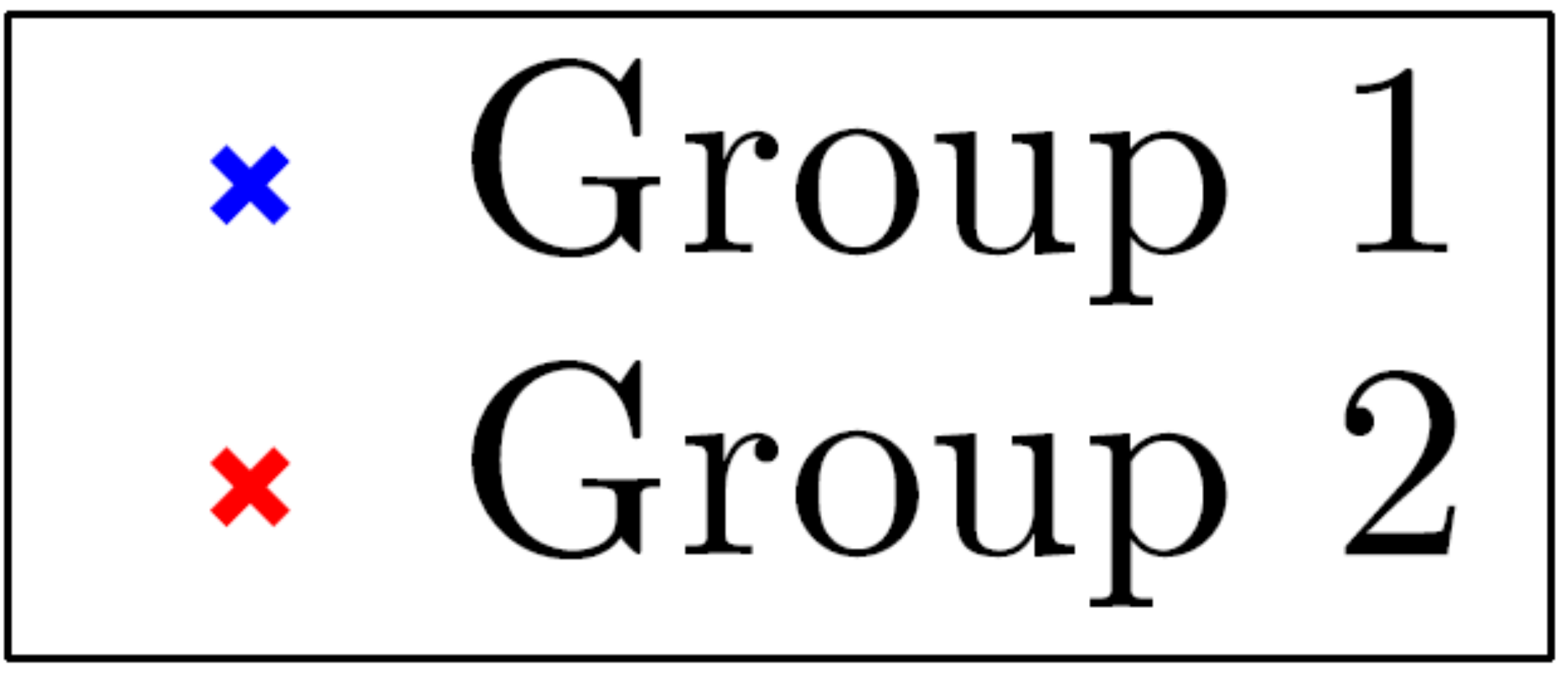}
  \caption{\small Scatter plots indicating that EMD-AMP (d) and, to a lesser extent, DFT-AMP (a) and DWPT-AMP (c) can be used to cluster data according to the time-frequency pattens most expressed. Results obtained by plotting the coefficient values outputted by each method (after classical MDS to two dimensions where appropriate) for 200 randomly selected time-series (100 from each group 1 (blue symbols) and 100 from group 2 (red symbols)). The average value for intermittency over all time-series in the set $\bar{\phi}$ is 0.8697. Parameter values for synthetic data generation as for Figure \ref{fig:beta_varphi}l except $\varphi=1.5$.}
 \label{fig:nonstationary_noise_scatter}
 \end{figure}

Scatter plots of the time-frequency features obtained using every AMP method considered in this work are shown in Figure \ref{fig:nonstationary_noise_scatter}. These have been symbolised based on whether the time-series are members of group 1 (blue symbols) or group 2 (red symbols). EMD-AMP (Figure \ref{fig:nonstationary_noise_scatter}d) clearly clusters the two groups according to the time-frequency patterns they most express. So do DFT-AMP (a) and DWPT-AMP (c), but not to the same extent. DWT-AMP (b) fails to cluster the data correctly in this instance. 

The success of EMD-AMP is related to the fact that its basis vectors permit a more parsimonious model of the data's underlying time-frequency patterns. Indeed for all cases considered in this section, applying EMD to the aggregate produces just two IMFs - each corresponding to one of the two intrinsic time-frequency patterns of the data. This is still true even when such patterns are non-stationary with time varying frequencies. 

In contrast, Fourier basis vectors (with their fixed frequencies) are incapable of succinctly modelling the intrinsic time-frequency patterns that exist in data with a frequency that varies over time. Similarly, the rigidity of the DWT means it produces wavelet vectors which individually only model an underlying time-frequency pattern for a small proportion of time \emph{and} for a small proportion of its frequency band. The basis vectors produced via the DWPT can only individually model either (i) a proportion of the frequency of an underlying pattern over the whole time domain, (ii) all of the underlying patterns frequency band but only for a short period of time, or (iii) neither. This means that no single basis vector obtained via the DFT, DWT or DWPT may individually capture the complete time-frequency patterns that underpin non-stationary data. Despite these weaknesses, these basis vectors are sufficiently similar to underlying time-frequency patterns within the data for DFT-AMP, DWT-AMP and DWPT-AMP to still yield reasonable results.

It is also notable that for stationary data (see the results in Figure \ref{fig:beta_varphi} (a,d,g,j)) the performances of EMD-AMP, DFT-AMP and DWPT-AMP are almost identical. In this instance, this is due to each method decomposing the aggregate into two almost identical components: one corresponding to the intrinsic oscillation with fixed period 2 and the other to the oscillation with fixed period 4.

Matlab code used to produce the synthetic data and obtain the results in this section is available at \newline https://github.com/duncan-barrack/AMP.
\newpage
\subsection{MIT reality mining data set}
\label{sec:real_world}
\noindent In order to provide evidence that AMP can be used to achieve meaningful results when applied to real world data we consider the MIT Reality Mining dataset. This set comprises event data pertaining to the times and dates at which MIT staff and students made a total of 54 440 mobile phone calls over a period from mid 2004 until early 2005. The average intermittency measure across the population is high ($\bar{\phi}=0.544$) and thus the data set is an excellent candidate for the AMP method. While no ground truth exists for this data, we utilise additional co-variates within the data set as a qualitative proxy for a ground truth for a useful segmentation. In particular, we use participants' affiliation (Media lab or Sloan business school, see \cite{eagle2009inferring} for details) as the proxy. Because of its excellent performance in the previous section, we choose to use the EMD-AMP variant.
\subsubsection{Results}
\label{sec:MIT_results}
\begin{figure}[htbp]
\centering
\hspace*{-0.8cm}
\includegraphics[width=0.55\textwidth]{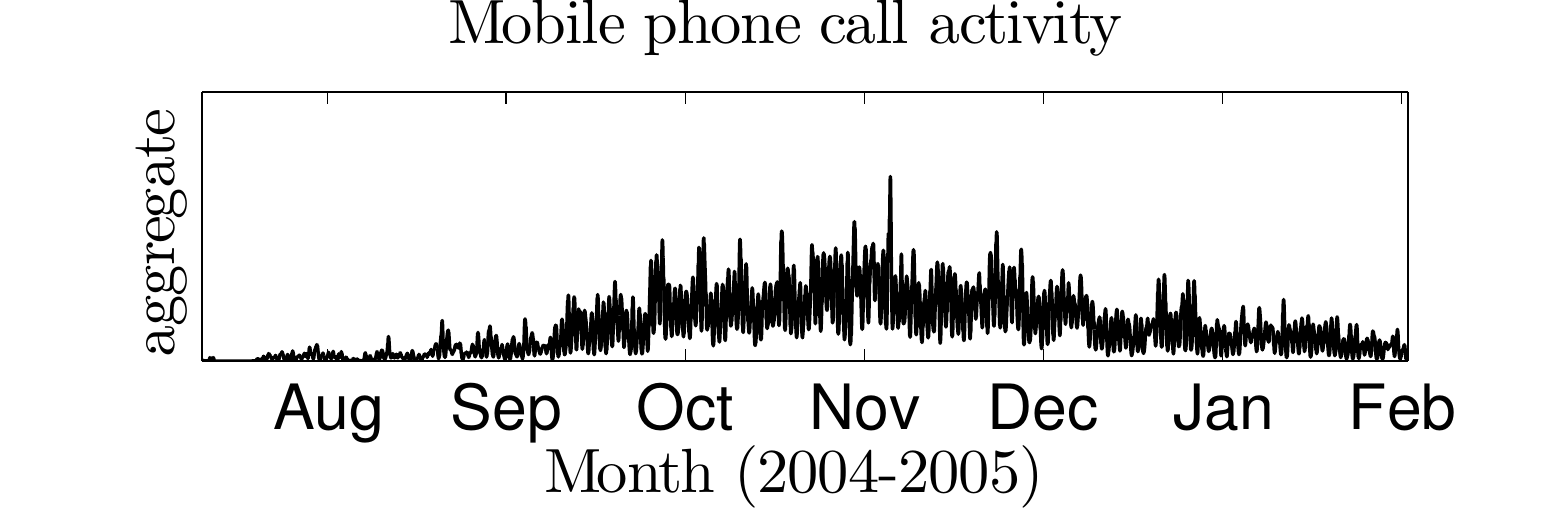}
\hspace*{-0.8cm}
\includegraphics[width=0.55\textwidth]{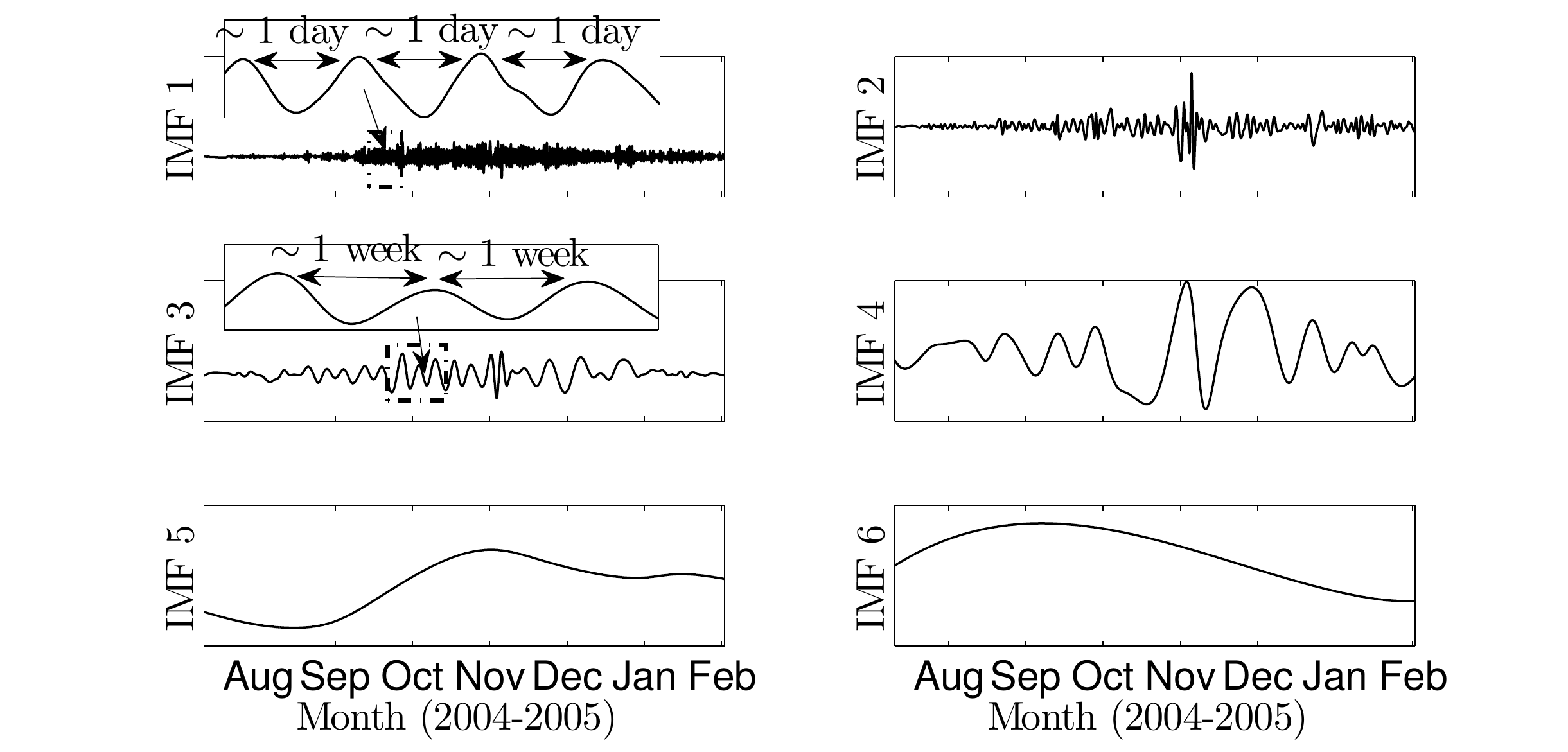}
\caption{\small Aggregate signal and decomposition obtained via EMD revealing six intrinsic time-frequency patters of the MIT communications data. IMF 1 has a period of 1 day and corresponds to the circadian cycle, while IMF 3 has a period of a week and corresponds to the seven day working week/weekend cycle.}
\label{fig:mit_sig_and_decomp}
\end{figure}

\noindent The aggregate and normalised IMFs of the MIT data are shown in Figure \ref{fig:mit_sig_and_decomp}. Interestingly, many of the IMFs have a physical interpretation. The first IMF has a period of almost exactly a day. This is most likely generated by the natural 24-hour circadian rhythm which will cause individuals to make a large proportion of their phone calls during the day and early evening. IMF 3 has a period of one week and most likely corresponds to the propensity of study participants to make more phone calls during the working week than at weekends. IMF 6 peaks in September/October before falling again in December/January. It is likely that this function corresponds to the changes in activity between the Fall term (September to December/January) and the holiday periods (over the summer and after Christmas) at MIT. 

\begin{figure}[htbt]
\centering
\includegraphics[width=0.23\textwidth]{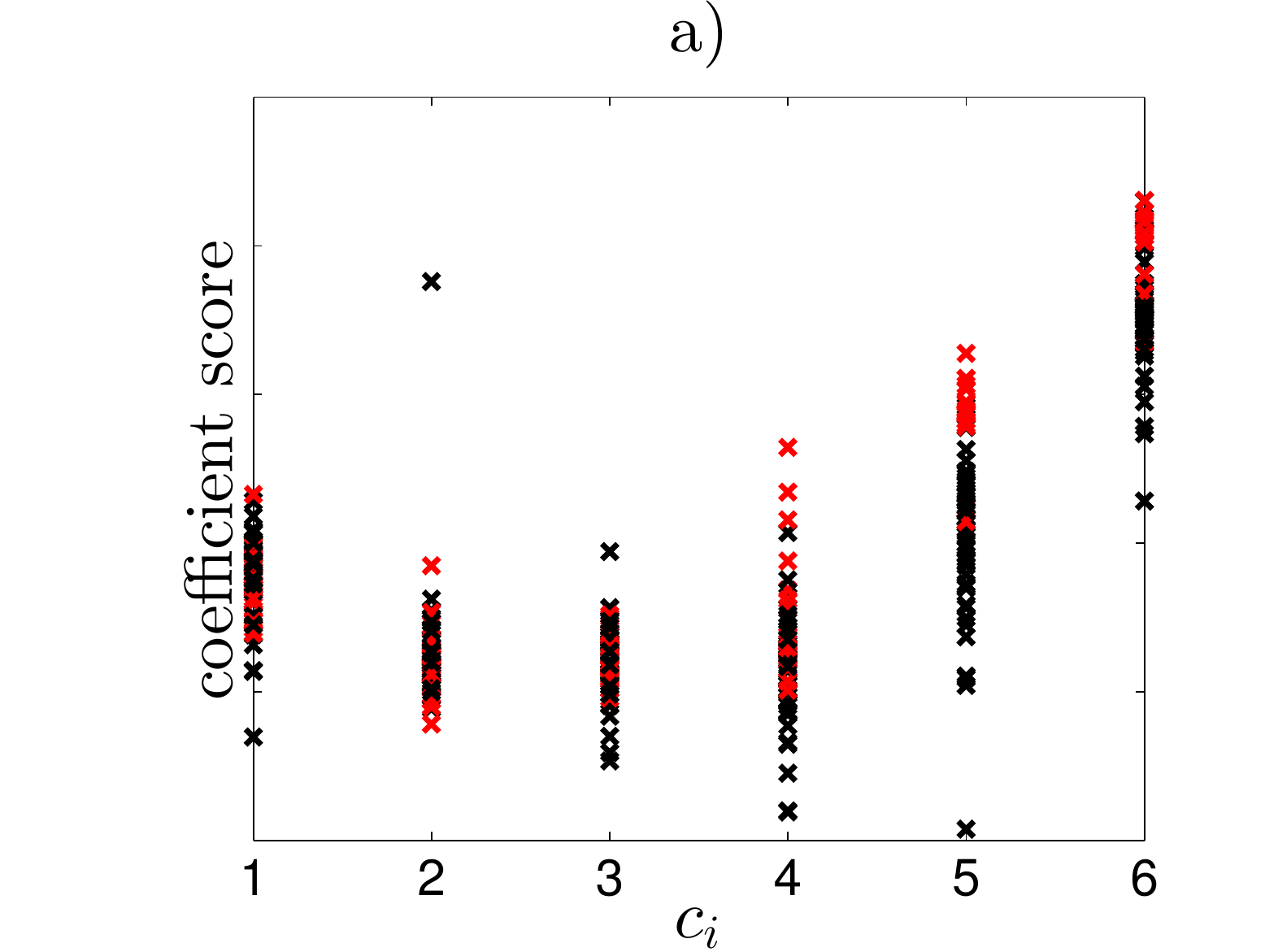}
\includegraphics[width=0.23\textwidth]{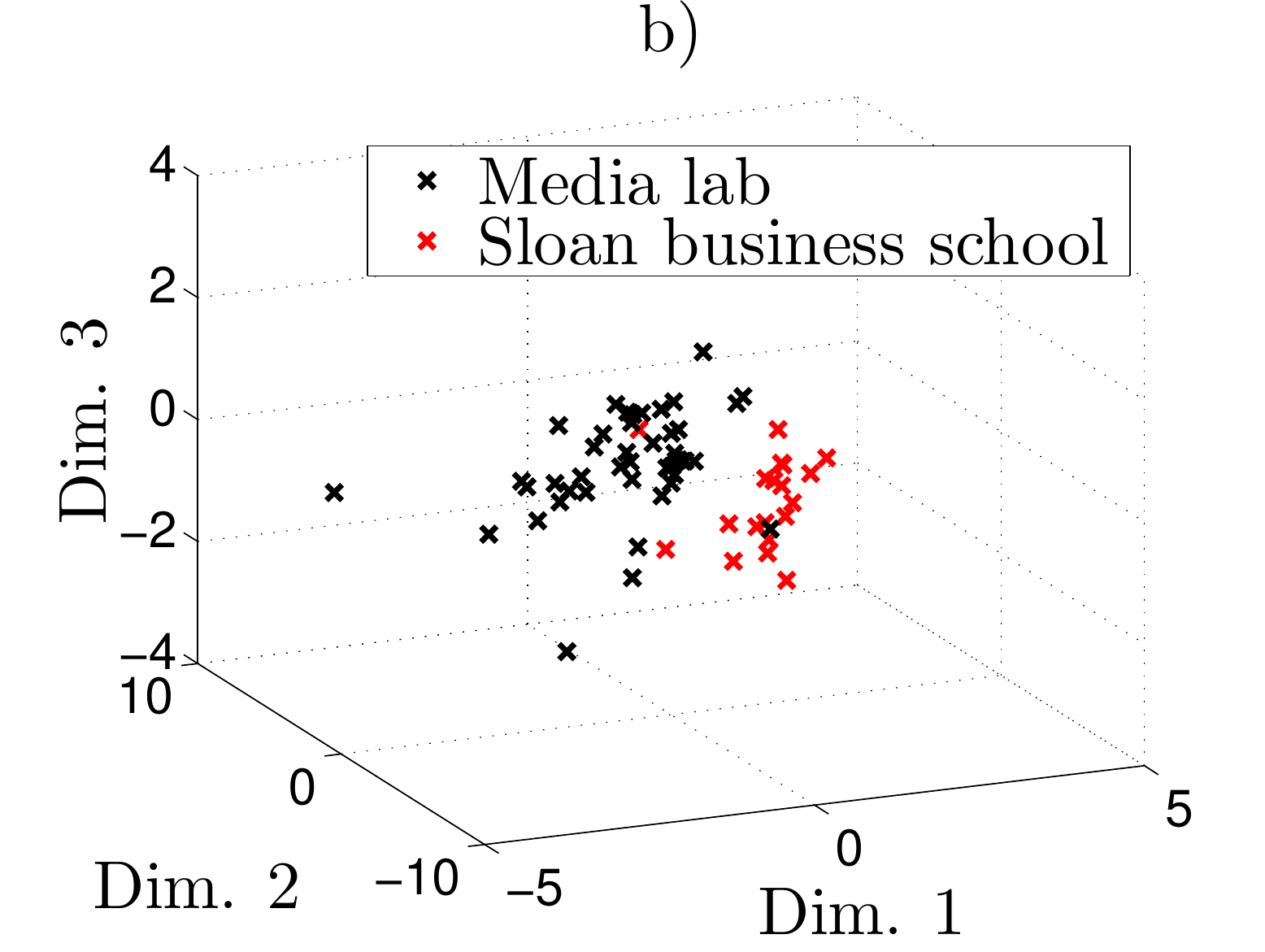}
\includegraphics[width=0.23\textwidth]{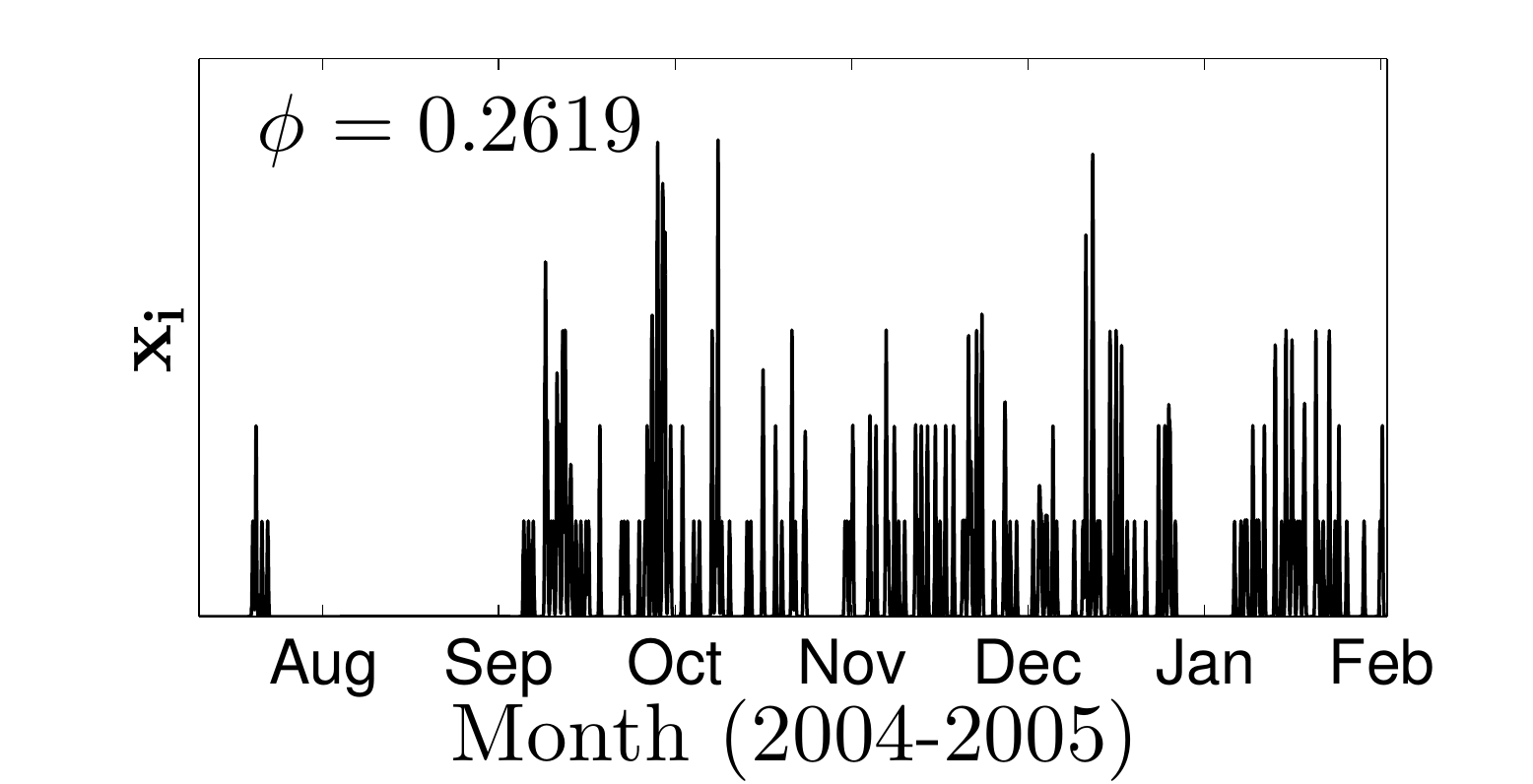}
\includegraphics[width=0.23\textwidth]{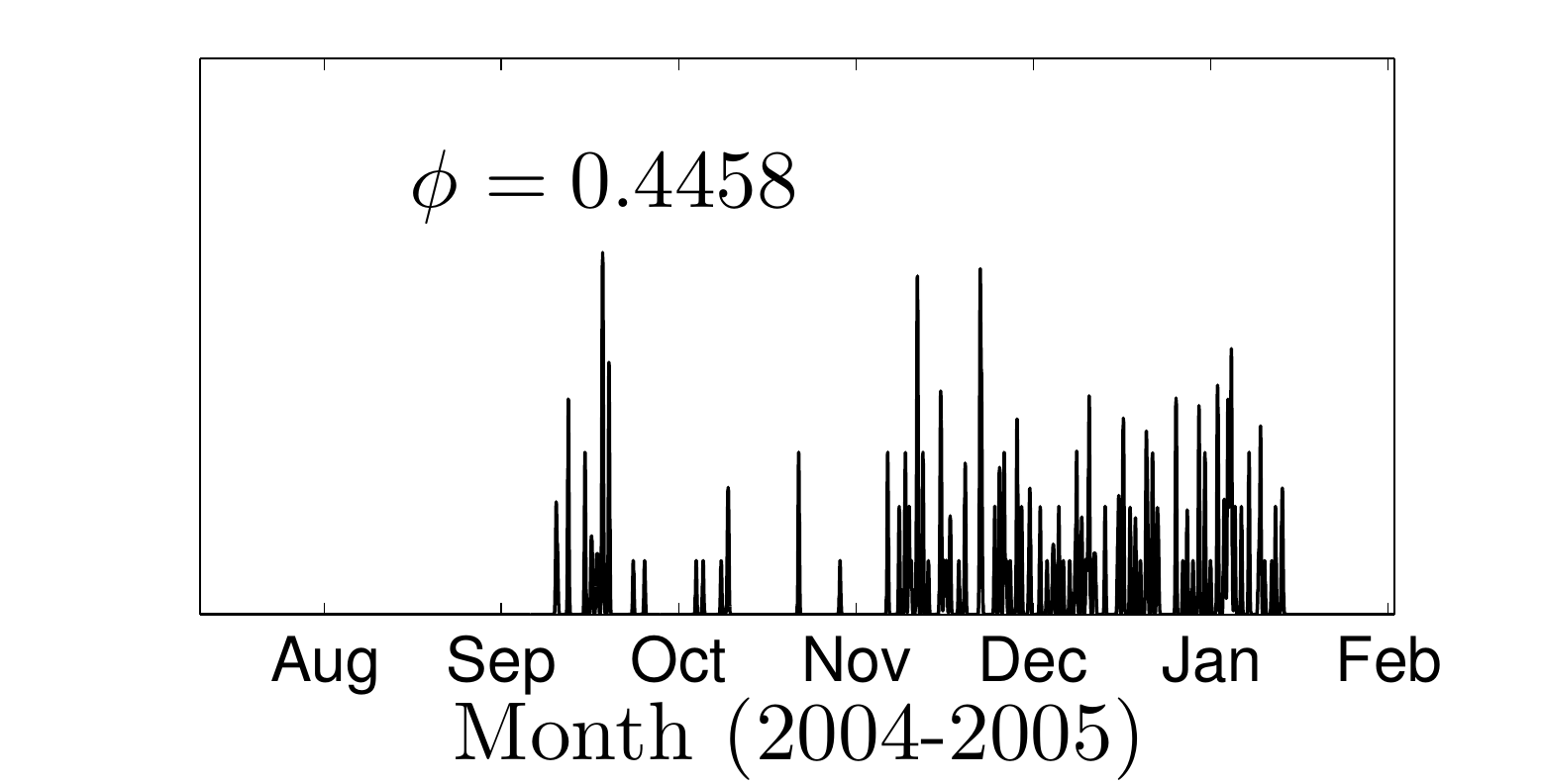}
\caption{\small Scatter plots indicating that Media lab and Sloan business school affiliates can be clustered according the the time-frequency patterns they most express. a) shows fitted coefficient values corresponding to each IMF. b) is a three dimensional representation of this data obtained via classical MDS. The bottom plots show the time-series of two study participants which are both clearly intermittent (intermittency measure values $\phi$ are given in the plots).}
\label{fig:mit_coef_dist}
\end{figure}

The extracted feature values for the 65 individuals who make the most phone calls are plotted in Figure \ref{fig:mit_coef_dist}, together with two clearly intermittent time-series of two randomly chosen individuals. The scatter plots have been symbolised based on whether the individuals were members of the reality mining group or the Sloan business school at MIT. Interestingly, from the three dimensional representation (Figure \ref{fig:mit_coef_dist}b) of the feature values, with the exception of a handful of individuals, the individuals from the two groups are separated from each other. Recall that the higher the feature value, the more the corresponding IMF is expressed in that individual's communications activity. From Figure \ref{fig:mit_coef_dist}a it can be seen that Sloan business school affiliates (red symbols) have, on average, larger coefficient values corresponding to IMFs 4-6 than the Media lab affiliates (black symbols). From this we can infer that the frequency patterns corresponding to IMFs 4-6 are expressed more strongly in the communication patterns of members of the Sloan business school.

\section{Conclusions and further work}
\label{sec:conclusions}
\noindent In this work we have addressed the problem of extracting pertinent features from intermittent time-series data containing \textcolor{black}{time-frequency patterns}. We have introduced a new approach entitled aggregation, mode decomposition and projection (AMP). The efficacy of AMP has been demonstrated by applying it to extensive synthetic data as well as to a real world communications data set with intermittent characteristics. \textcolor{black}{From a practical perspective AMP also holds promise due to its computational cheapness (for example, The results in Section \ref{sec:MIT_results} took less than two minutes to calculate using Matlab on a PC utilising single 3.2 GHz processor)}. We note that even though our intermittence measure (equation (\ref {eq:aggregate_time})) may not capture the degree of intermittence of all types of intermittent data, AMP will still be effective in these situations.

In terms of further work, it would be interesting to compare the performance of AMP clustering to techniques not considered in this paper such as a clustering approach underpinned by the Lomb-Scargle Periodogram \cite{lomb1976least} which was developed specifically for irregularly sampled data. Another time-domain based clustering method which we did not consider is the recently proposed shapelet based clustering approach \cite{mueen2012clustering} where `local patterns' in the data are exploited. However, as the values of intermittent times-series fluctuate only briefly from the the modal value, data rarely displays meaningful local patterns and our intuition is that shapelet based clustering is not appropriate in this instance. A comparison with model based approaches such those underpinned by switching Kalman filters \cite{murphy1998switching} would also be interesting.

\textcolor{black}{The focus of this paper's evaluation} has been on using AMP derived features for clustering. However, AMP features are also suited to the classification task. There is huge scope to apply an AMP based clustering and/or classification approach to many other types of intermittent time-series data and this would provide an interesting avenue for future work. For example, retail transaction data is characterised by the sporadic activity of customers who make a small number of purchases over a long period of time. The timings of these purchases will be dictated by time-frequency patterns corresponding to human behavioural patterns such as the 24 hour circadian rhythm, or 7 day working week/weekend. Another example is the number of trips made by an individual on public transport. These are unlikely to total more than a handful per day, but the timings of these trips is likely to follow time-frequency patterns corresponding to the traveller's commuting behaviour or leisure plans.

\section{Acknowledgements}
\noindent This work was funded by EPSRC grant EP/G065802/1 - Horizon: Digital Economy Hub at the University of Nottingham and EPSRC grant EP/L021080/1 - Neo-demographics: Opening Developing World Markets by Using Personal Data and Collaboration. 

%
\bibliographystyle{abbrv}
\bibliography{barrack}  
%
%

\end{document}